%% file: aaai24.tex
\documentclass[letterpaper,dvipsnames]{article} % DO NOT CHANGE THIS
\usepackage{aaai24}  % DO NOT CHANGE THIS
\usepackage{times}  % DO NOT CHANGE THIS
\usepackage{helvet}  % DO NOT CHANGE THIS
\usepackage{courier}  % DO NOT CHANGE THIS
\usepackage[hyphens]{url}  % DO NOT CHANGE THIS
\usepackage{graphicx} % DO NOT CHANGE THIS
\usepackage{natbib}  % DO NOT CHANGE THIS AND DO NOT ADD ANY OPTIONS TO IT
\usepackage{caption} % DO NOT CHANGE THIS AND DO NOT ADD ANY OPTIONS TO IT
\usepackage{algorithm}
\usepackage{algorithmic}
\usepackage{xspace}

\usepackage{tablefootnote}
\usepackage{color, xcolor}
\usepackage{tabularx}

\definecolor{red1}{rgb}{1, 0.6, 0.6}

\newcommand{\method}{\texttt{CELLO}\xspace}
\newcommand{\num}{523\xspace}
\newcommand{\modelNum}{34\xspace}
\newcommand{\modelNumch}{19\xspace}
\newcommand{\modelNumeg}{15\xspace}

\usepackage{booktabs}
\usepackage{multicol}
\usepackage{multirow, makecell, caption}
\usepackage{colortbl}
\usepackage{tikz}
\usepackage{arydshln}
\usepackage{pgfplots}
\usepackage{amsmath}
\usepackage{amssymb}

\makeatletter
\def\adl@drawiv#1#2#3{%
        \hskip.5\tabcolsep
        \xleaders#3{#2.5\@tempdimb #1{1}#2.5\@tempdimb}%
                #2\z@ plus1fil minus1fil\relax
        \hskip.5\tabcolsep}
\newcommand{\cdashlinelr}[1]{%
  \noalign{\vskip\aboverulesep
           \global\let\@dashdrawstore\adl@draw
           \global\let\adl@draw\adl@drawiv}
  \cdashline{#1}
  \noalign{\global\let\adl@draw\@dashdrawstore
           \vskip\belowrulesep}}
\makeatother

\usepackage{newfloat}
\usepackage{listings}
\DeclareCaptionStyle{ruled}{labelfont=normalfont,labelsep=colon,strut=off} % DO NOT CHANGE THIS
\floatstyle{ruled}
\newfloat{listing}{tb}{lst}{}
\floatname{listing}{Listing}
\title{Can Large Language Models Understand Real-World Complex Instructions?}

\author{
    Qianyu He\textsuperscript{\rm 1}, 
    Jie Zeng\textsuperscript{\rm 1},
    Wenhao Huang\textsuperscript{\rm 1},
    Lina Chen\textsuperscript{\rm 2},
    Jin Xiao\textsuperscript{\rm 2},
    Qianxi He\textsuperscript{\rm 1},
    Xunzhe Zhou\textsuperscript{\rm 1},\\
    Lida Chen\textsuperscript{\rm 1},
    Xintao Wang\textsuperscript{\rm 1},
    Yuncheng Huang\textsuperscript{\rm 1},
    Haoning Ye\textsuperscript{\rm 1},
    Zihan Li\textsuperscript{\rm 1},\\
    Shisong Chen\textsuperscript{\rm 4},
    Yikai Zhang\textsuperscript{\rm 1},
    Zhouhong Gu\textsuperscript{\rm 1},
    Jiaqing Liang\textsuperscript{\rm 2}\thanks{Corresponding author.},
    Yanghua Xiao\textsuperscript{\rm 1,3}\footnotemark[1]
}
\affiliations{
    \textsuperscript{\rm 1}Shanghai Key Laboratory of Data Science, School of Computer Science, Fudan University\\
    \textsuperscript{\rm 2}School of Data Science, Fudan University\\
     \textsuperscript{\rm 3}Fudan-Aishu Cognitive Intelligence Joint Research Center, Shanghai, China\\
       \textsuperscript{\rm 4}Shanghai Institute of AI for Education and School of Computer Science and Technology, East China Normal University\\
    \{qyhe21, jzeng23, whhuang21, lnchen23, jinxiao23, qxhe23, chenld23, xtwang21, yunchenghuang22, zihanli21, ykzhang22, zhgu22\}@m.fudan.edu.cn, sschen@stu.ecnu.edu.cn, \{hnye19, xzzhou20, liangjiaqing, shawyh\}@fudan.edu.cn
}

\usepackage{bibentry}
\begin{document}

\maketitle

\begin{abstract}

\input{000abstract}
\end{abstract}

\section{Introduction}
\input{010intro}

\section{Related Work}
\input{020related}

\section{\method Benchmark}
\input{030benchmark}

\input{040experiment}

\section{Conclusion}
\input{050conclusion}

% \clearpage
\bibliography{aaai24}
\input{100appendix}

\end{document}

%% file: 000abstract.tex
Large language models (LLMs) can understand human instructions, showing their potential for pragmatic applications beyond traditional NLP tasks. However, they still struggle with complex instructions, which can be either complex task descriptions that require multiple tasks and constraints, or complex input that contains long context, noise, heterogeneous information and multi-turn format. Due to these features, LLMs often ignore semantic constraints from task descriptions, generate incorrect formats, violate length or sample count constraints, and be unfaithful to the input text. Existing benchmarks are insufficient to assess LLMs’ ability to understand complex instructions, as they are close-ended and simple. To bridge this gap, we propose \method, a benchmark for evaluating LLMs' ability to follow complex instructions systematically. We design eight features for complex instructions and construct a comprehensive evaluation dataset from real-world scenarios. We also establish four criteria and develop corresponding metrics, as current ones are inadequate, biased or too strict and coarse-grained. We compare the performance of representative Chinese-oriented and English-oriented models in following complex instructions through extensive experiments. 
Resources of \method are publicly available at \url{https://github.com/Abbey4799/CELLO}.

%% file: 010intro.tex
The emergence of large-scale models~\cite{brown2020language, chowdhery2022palm, touvron2023llama} has yielded noteworthy transformations in real-world applications~\cite{richards2023auto, liu2023agentbench}.
These models are able to understand a wide range of human instructions, spanning from casual conversations~\cite{taori2023stanford} to complex problems solving~\cite{brown2020language}.
Since human instructions are massive and diverse, traditional academic benchmarks that focus on specific tasks are no longer sufficient to evaluate LLMs~\cite{zhong2023agieval, chia2023instructeval}.

\input{figs/010intro.tex}
\input{figs/030framework.tex}

% 复杂指令的理解对模型而言是个有挑战的事

% 真实场景中有很多复杂指令，且不是传统的Benchmark中简单指令
Real-world applications often involve a diverse range of complex instructions that significantly differ from the simple and common instructions in current benchmarks~\cite{hendrycks2020measuring, huang2023c}, as shown in Fig.~\ref{fig:010intro}.
% In general, complex instructions understanding remains quite challenging for LLMs~\cite{xu2023wizardlm, luo2023wizardcoder, zhou2023lima}.
Instruction generally consists of two parts~\cite{honovich2022unnatural}: \textit{Task description} (mandatory) describes the task goal and \textit{Input text} (optional) provides reference texts for the model to answer questions or the history of multi-turn conversations, as shown in Fig.~\ref{fig:010intro}.
Hence, there can be two categories of complex instructions: \textit{complex task descriptions} and \textit{complex input}.
Regarding \textit{complex task descriptions}, models need to undertake multiple tasks (i.e. \underline{multi-tasking)} and there can be diverse restrictions describing the task, including \underline{semantics constraints} (e.g. the inclusion of key elements~\cite{zhou2023lima} or the use of predefined callable functions~\cite{liu2023agentbench}), \underline{format constraints} (e.g. the predefined format in few-shot scenarios~\cite{yao2023react} or structured format imitating human reasoning processes~\cite{liu2023agentbench}), \underline{quantity constraints} (e.g. word, sentence, or sample count regulating the length of model output~\cite{zhou2023controlled, yao2023collie}).
Regarding \textit{complex input}, the input text generally have \underline{long context}~\cite{an2023eval, liu2023lost}, \underline{noise} (e.g. colloquial expressions~\cite{guo2023close} and error accumulation caused by pipeline method~\cite{sun2023chatgpt}), \underline{heterogeneous information} (e.g. a combination of structured and unstructured data~\cite{zha2023tablegpt}), and in the form of \underline{multi-turn}~\cite{ding2023enhancing}.

% 现实应用指令的复杂性attribute to account for many existing errors of LLMs, posing challenges for LLMs
The complexity of real-world instructions accounts for prevalent errors observed in LLMs.
% Owing to the above features of complex instructions also depicted in Fig.~\ref{fig:030framework}, LLMs often encounter \textit{four} types of errors during complex instruction understanding.
As shown in Fig.~\ref{fig:010intro}, LLMs may (1) ignore \underline{semantic constraints} from task description(s)~\cite{zhou2023lima}, (2) generate answers in incorrect \underline{format}~\cite{qin2023toolllm}, or (3) violate the length or sample \underline{count constraints}~\cite{zhou2023controlled}, especially when \underline{multiple tasks} are required to be performed.
Moreover, models can (4) be unfaithful to the input text, especially when it is \underline{long}, \underline{noisy}, \underline{heterogeneous} or in the form of \underline{multi-turn}~\cite{li2023halueval, an2023eval}.
Overall, complex instructions pose challenges to current models.

However, existing benchmarks are insufficient for effectively assessing the ability of LLMs to understand complex instructions.
On one hand, Fig.~\ref{fig:010intro} shows that existing benchmarks are either close-ended~\cite{huang2023c, zhong2023agieval, yu2023kola} or contain common and simple instructions~\cite{srivastava2023beyond, chia2023instructeval, dubois2023alpacafarm}, which fail to mirror the complexity of real-world instructions.
On the other hand, even though certain benchmarks cover some of the above features of complex instructions, such as count restriction~\cite{zhou2023controlled, yao2023collie}, semantic restriction~\cite{chen2022controllable}, and long text understanding~\cite{an2023eval}, they only encompass isolated features, while real-world instructions comprehensively cover these features~\cite{zhou2023lima}.
Overall, none of the existing benchmarks systematically study the complex instructions understanding ability of LLMs.

In this paper, we propose \method, a benchmark for evaluating the \textbf{\underline{C}}ompl\textbf{\underline{E}}x instruction understanding ability of \textbf{\underline{L}}arge \textbf{\underline{L}}anguage M\textbf{\underline{O}}dels systematically.
% a benchmark for evaluating LLMs' ability to follow complex instructions systematically.
The framework of our benchmark is shown in Fig.~\ref{fig:030framework}.
% 稍微解释一下close-ended，去掉那么多citation
% 我们从八个维度刻画复杂指令，建framework to复杂指令的框架with eight features
% Accordingly, we propose a novel evaluation system. 再说为什么现有的benchmark不足
% citation 空格 binary pass rate，strict and coarse-grained太vague了
% propose criteria
% first design eight features for complex instructions and construct an evaluation dataset covering these features comprehensively.
As existing benchmarks only cover isolated features of complex instructions, we establish a comprehensive framework comprising eight features of complex instructions.
Accordingly, we propose a novel evaluation system comprised of four criteria along with their corresponding metrics.
The current evaluation criteria are insufficient to comprehensively reflect the ability of LLMs to understand complex instructions for the following reasons.
First, complex instructions in real-world scenarios are open-ended~\cite{xu2023wizardlm}, thus the criteria commonly used for close-ended benchmarks are not suitable in such cases~\cite{hendrycks2020measuring}.
Moreover, many studies adopt GPT4 evaluation for automated open-ended assessment, which introduces bias problems \cite{wang2023large}.
Furthermore, the binary pass rate adopted by the benchmarks containing complex instructions is strict and coarse-grained, resulting in universally low scores for smaller LLM without discrimination~\cite{liu2023agentbench, qin2023toolllm}.
% Hence, we establish four criteria based on the four types of errors that LLMs might face while comprehending instruction, as previously introduced, and develop corresponding metrics to evaluate them automatically.

Overall, our contributions are mainly four-fold: 
\begin{itemize}
    \item To the best of our knowledge, we are the first to systematically investigate the ability of LLMs to follow complex instructions. 
We propose a comprehensive set of features for complex instructions, facilitating both dataset construction and evaluation criteria design.
    \item We construct a complex instruction dataset from real-world scenarios, containing \num samples encompassing nine tasks, effectively covering our specified features.
    Specifically, we propose a two-stage framework for constructing the evaluation dataset for LLM's complex instruction understanding.
    \item We design four evaluation criteria and corresponding automatic metrics for assessing LLMs' ability to understand complex instructions in a comprehensive and discriminative way.
    % 分析了不同主导语言的模型对中文复杂指令的理解能力
    \item We compare \modelNumch representative Chinese-oriented models and \modelNumeg representative English-oriented models' performance on our benchmark.
\end{itemize}

%% file: figs/010intro.tex
\begin{figure}[t] 
    \centering
        \includegraphics[width=1\linewidth]{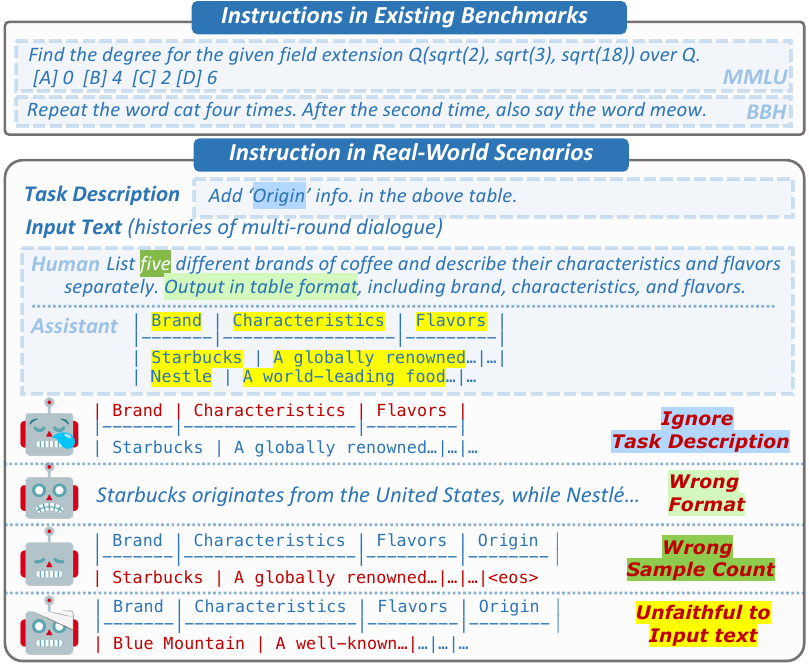} 
    \captionsetup{font={small}} 
    \caption{Existing benchmarks generally contain simple and common instructions. However, the complex instructions in real-world scenarios are a composition of multiple features, such as constraints on the output format, number of output samples, key elements of the output, and heterogeneity of input texts in the given example. The understanding of complex instructions poses challenges to current models.}
    \label{fig:010intro}
\end{figure}

% 真实场景中的复杂指令往往是多个特征的结合，例如：输出形式约束、输出样本数约束、输出的Key elements约束、异构的输入文本。

%% file: figs/030framework.tex
\begin{figure*}[t] 
    \centering
        \includegraphics[width=1\linewidth]{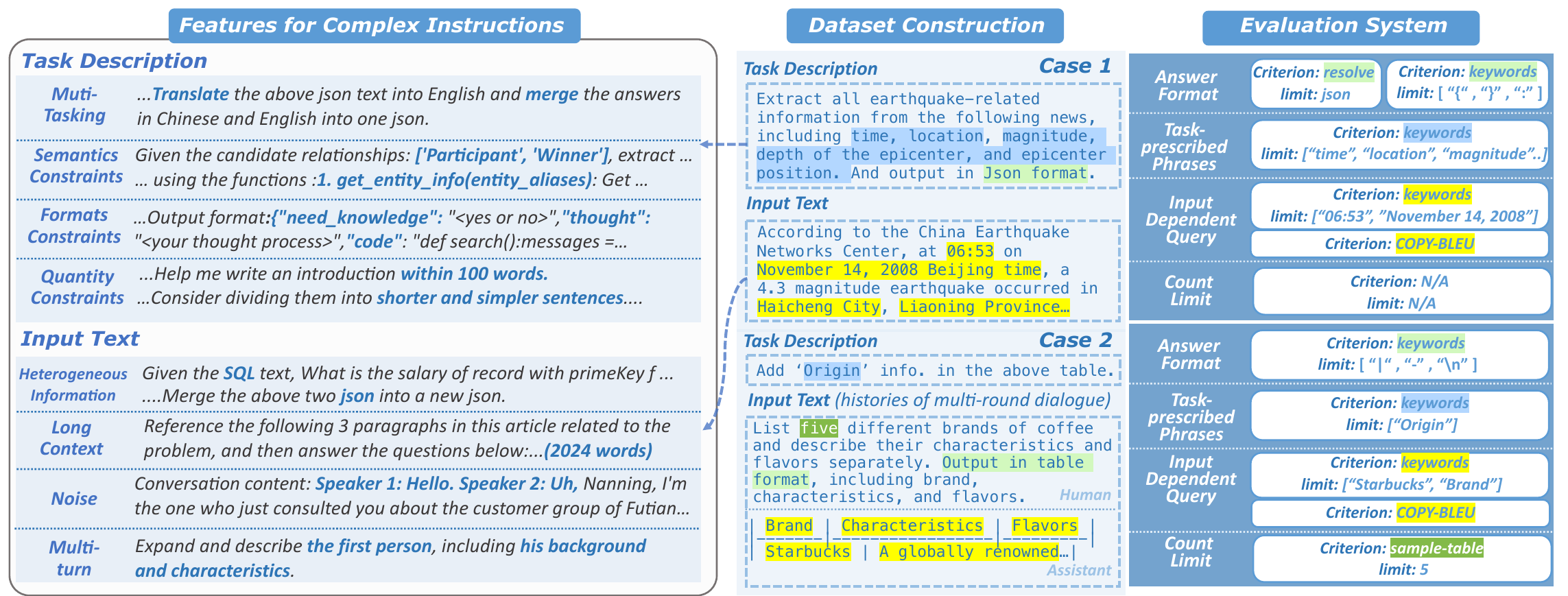} 
    \captionsetup{font={small}} 
    \caption{The framework of our benchmark design. We first establish a framework containing eight features for complex instructions, then construct an evaluation dataset covering nine tasks, and finally propose four evaluation criteria along with their corresponding metrics.}
    \label{fig:030framework}
\end{figure*}

%

%% file: 020related.tex
\paragraph*{Evaluation for LLMs}
% Benchmark: 知识、复杂推理、代码、工具
% GPT4 evaluation
% Arena

% 很多Benchmarks从多个维度衡量大模型。一些Benchmark（HELM, INSTRUCTEVAL, AlpacaEval）划分了全面的评估维度并整合了现有评估数据集，包含对知识（MMLU）、推理、有害性评估等等。
% 特别地，主流的benchmark集中于对模型的知识、编程能力、复杂推理的评估，在它们的Benchmark上，小模型要么甚至超过了ChatGPT等大模型，要么远远低于ChatGPT以至于没有区分度。
% 最近，很多Benchmark关注模型在特别方面能力，比如工具调用、Agent、长文本处理等。
% 对于自动评估标准而言，对于close-ended或者可以根据成功与否判断的任务，研究者们用客观的评估标准，比如acc、f1、success rate。对于open-ended任务而言，研究者们普遍采用GPT4 evaluation的方法，但也有人指出GPT4 evaluation存在bias。
% 整体而言，这些Benchmark存在以下局限性：（1）Instruction偏简单，不能反应真实场景下用户的复杂需求（2）任务太难并且评估标准太粗粒度，导致对复杂任务而言小模型没有区分度（3）对于Open-ended任务而言，自动化评估存在bias。
Many benchmarks propose comprehensive evaluation frameworks that integrate existing evaluation datasets~\cite{liang2022holistic, zhong2023agieval, dubois2023alpacafarm, chia2023instructeval}.
Mainstream benchmarks primarily focus on assessing knowledge~\cite{huang2023c, gu2023xiezhi, yu2023kola}, programming~\cite{chen2021evaluating}, and complex reasoning~\cite{cobbe2021training, srivastava2023beyond}.
Recently, many benchmarks focus on specific capabilities of models, such as tool utilization~\cite{qin2023toolllm}, acting as agents~\cite{liu2023agentbench}, and handling long texts~\cite{an2023eval}.
However, none of the existing benchmarks systematically investigate the ability of LLMs to follow complex instructions.
Their evaluation criteria have several limitations when evaluating complex instruction understanding.
First, the close-ended benchmarks fail to mirror the complexity of the real-world instructions~\cite{huang2023c, gu2023xiezhi, zhong2023agieval}.
Also, the binary success rate~\cite{chen2021evaluating, qin2023toolllm, liu2023agentbench} is too strict and coarse-grained, resulting in weak discrimination.
Moreover, GPT-4 automatic scoring introduces bias problems~\cite{wang2023large}.
Overall, the existing benchmarks and their criteria are insufficient to effectively assess LLMs' ability to understand complex instructions.
% systematically investigate LLMs' ability to understand complex instructions.
% automated evaluation criteria, objective metrics such as accuracy, F1 score~\cite{zhong2023agieval, yu2023kola}, and binary success rate~\cite{chen2021evaluating, qin2023toolllm, liu2023agentbench} are commonly employed for close-ended or pass-or-fail tasks.
% While for open-ended tasks, researchers generally adopt GPT4 as the evaluator~\cite{an2023eval}.
% On one hand, the instructions in current benchmarks are either short and simplistic or only evolve isolated features of complex instructions, thus adequately reflecting the complexities of real-life scenarios.
% On the other hand, the existing benchmarks incorporate multiple features of complex instructions but adopt strict and coarse-grained metrics making it difficult to discriminate between smaller models.
% the instructions are generally short and simplistic thus adequately reflecting the complexities of real-life scenarios~\cite{xu2023wizardlm}. 
% Also, for the tasks evolving, the evaluation criteria are too coarse-grained, resulting in minimal distinction among smaller LLMs~\cite{qin2023toolllm, liu2023agentbench, chen2021evaluating}.
% Lastly, the automated scoring by GPT-4 introduces issues such as instability and bias~\cite{wang2023large}.

\paragraph*{Complex Instruction Following}
% WizardLM
% WizardCoder
% LIMA
% Orca: Progressive Learning from Complex Explanation Traces of GPT-4

% 海量的instruction tuning数据普遍存在任务单一、指令简单的情况，这导致它们难以理解真实场景中存在的复杂指令。不同的工作采用不同的方法复杂化指令数据从而提升模型理解指令理解的能力。WizardLM，WizardCoder提出六类策略改进self-instruct，基于少量的手写种子数据生成了大量复杂指令。LIMA采用众包让人类搜集了少量高质量且复杂的用户query-response对。Orca设计prompt诱导GPT4对简单指令生成推理步骤从而复杂化训练数据。然而，缺乏一个Benchmark系统地分析模型对复杂指令的理解能力。
The current datasets generally have simple and common instructions, making LLMs challenging to follow complex instructions in real-world scenarios~\cite{zhou2023lima, xu2023wizardlm}.
Various methods have been proposed to improve models' understanding of complex instructions.
~\citet{xu2023wizardlm, luo2023wizardcoder} propose six strategies to generate complex instructions based on a small set of handwritten seed data. 
~\citet{zhou2023lima} utilizes crowdsourcing to collect a limited number of high-quality and complex user query-response pairs. 
~\citet{mukherjee2023orca} induce GPT4 to generate reasoning steps for simple instructions, thereby complexifying the training data. 
Despite the advancements, there is a lack of a benchmark for systematically evaluating models' understanding of complex instructions.

\paragraph*{Evaluation for Constrained Instructions}
% CommonGen: A Constrained Text Generation Challenge for Generative Commonsense Reasoning侧重常识的正确性
% Controllable Text Generation with Language Constraints侧重topic相关的constraints
% COLLIE: Systematic Construction of Constrained Text Generation Tasks侧重Grammar相关的constraints，不太符合真实场景
% Controlled Text Generation with Natural Language Instructions也定义了很多constraints，不太符合真实场景

% 传统对Text Generation的约束可分为硬约束（structural/lexical）和软约束（semantic）。regarding automatic metrics, 对于软约束，常用classifier-based的方法去判断。对于硬约束，常用rule-based判断，例如coverage和success rate。

% 不少工作系统性研究LLMs理解constrainted instruction的能力。
% COLLIE提出了一个基于grammar的约束生成框架，主要考虑词数、位置两类lexical的限制。
% INSTRUCTCTG基于设计的五类约束自动化构建大规模带约束的文本生成训练数据。
% COGNACGEN是一个benchmark，对生成文本的话题进行限制的同时，还包含对需要规避的内容的限制。
% 然而，这些Benchmark的指令都过于简单，这些Benchmark涉及到的constraints也太过单一。
Many studies investigate the ability of LLMs to understand constrained instructions.
~\citet{yao2023collie} proposes a grammar-based framework for generating instructions with lexical constraints related to word count and position.
~\citet{zhou2023controlled} adopts five types of constraints to automatically construct large-scale constrained instructions.
~\citet{chen2022controllable} limits the topics of generated text while also including constraints on the content to be avoided.
However, the instructions of these benchmarks are simplistic, and the constraints they involve are narrow.

%% file: 030benchmark.tex
% \subsection{Complex Instructions}
% 我们定义Instruction由任务描述（必选）与输入文本（可选）组成。因此，在我们的基准测试集中包含两类复杂指令（1）任务描述本身有多种限制（2）输入文本很长、很多噪音。
% As shown in Fig.~\ref{}, \textit{Instruction} is defined as consisting of \textit{task description} (mandatory) and \textit{input text} (optional)~\cite{}. 
% % 我们的Benchmark中包含的Complex Instruction有以下特点：
% Hence, our benchmark contains two types of complex instructions with
% (1) multiple constraints in the task description, or (2) long, noisy, or structured input texts.

% 形式、约束、输出、异构、任务的混杂
% 对应现有工作的缺陷4
% 真实例子
% \hqy{items more conclude}

% As shown in Fig.~\ref{}, 在我们的Benchmark中，指令由以下两个部分组成：（1）任务描述（必选）：描述任务目标，通常包含对输出的内容、形式、可调用函数等具体细节的要求。（2）输入文本（可选）：包含模型回答问题时需要参考的文本、以及多轮对话的上下文历史记录。整体而言，我们Benchmark中包含指令的复杂性体现在以下几个维度，就任务描述而言：（1）输出形式多样：除了可解析的形式（例如json、xml、markdown表格等）还包括few-shot的样本格式示例（2）约束输出信息的内容：包含lexical constraints、可调用的函数等（3）约束输出信息的长度及数量：例如字数、句子数、样本数（4）多种任务混杂；就输入文本而言：（5）输入形式异构（6）输入文本有噪音：有无关或者错误信息的干扰（7）输入文本很长：由于我们的数据很多是从真实场景中搜集得到，因此as shown in 表~\ref可知我们benchmark中的数据普遍较长
As shown in Fig.~\ref{fig:030framework}, we first establish a framework containing eight features for complex instructions, then construct an evaluation dataset, and finally propose four evaluation criteria along with their corresponding metrics.
% As shown in Fig.~\ref{fig:030framework}, the instructions in our benchmark consist of two parts: \textit{Task description} (mandatory) and \textit{Input text} (optional).
%~\cite{honovich2022unnatural}
% (1) \textbf{Task description} (mandatory) describes the task goal, usually with detailed requirements, such as keywords, format, and callable functions.
% (2) \textbf{Input text} (optional) provides reference texts for the model to answer questions or the history of multi-turn conversations.
% Overall, the complex instructions have the various features as shown in Fig.~\ref{fig:030framework}.
% regarding the task description:
% (1) \textit{Output formats}, including parsable formats and few-shot sample format examples,
% (2) \textit{Output semantics}, such as lexical constraints, callable functions, and primary objects,
% (3) \textit{Output counts}, constraints on output length and sample quantity,
% (4) \textit{Multi-tasking}; 
% regarding the input text:
% (1) \textit{Heterogeneity}, (2) \textit{Length}, (3) \textit{Noise}, (4) \textit{Multi-turn}.

% % \input{tables/030examples}

\subsection{Dataset Construction}
% We construct an evaluation dataset to assess LLMs' understanding of complex instructions.
%We propose a two-stage framework for dataset construction.
We first collect data from real scenarios, covering 9 tasks.
Then we diversify the collected complex instructions through \textit{In-breadth Evolution} and complicate the collected simple instructions through \textit{In-breadth Evolution}.

\input{tables/030statistics}

\subsubsection{Data Source and Selected Tasks}
% \input{tables/030examples}
% 根据定义的硬约束，我们构建了评估大模型理解带约束的复杂指令的的数据集。整体而言，为了保证评估数据来源于真实场景的需求且防止数据泄漏的隐患，我们的数据有以下来源：
% （1）实际项目需求：ChatPDF、新闻和工业等领域的复杂抽取、录音摘要、智慧教材等一系列和企业合作的项目。
% （2）CuteGPT的用户使用的logs：搜集了四个月内的用户query log，去重后有9000条用户数据。从中，我们通过长度筛选、正则表达式匹配的方式挖掘带有硬性约束的复杂指令数据。
% The instructions within our benchmark are a composition of our defined features.
When constructing the dataset, we take into account its \textit{coverage} and \textit{representativeness}.
Regarding \textit{coverage}, we include common NLP tasks found in existing benchmarks~\cite{liang2022holistic}, while incorporating instructions with more complex task descriptions or input beyond those benchmarks. 
Moreover, we introduce specific tasks involving complex instructions, which align with common real-world applications for LLMs.
Regarding \textit{representativeness}, instructions are gathered from 90,000 user interaction logs over six months with our implemented chatbot.
% To ensure that the evaluation data comes from real scenarios and to prevent the risk of data leakage, our data has the following sources: 
% (1) \textbf{Real-world project requirements}, a series of projects in cooperation with enterprises, such as ChatPDF, industry information extraction, and audio summarization, etc. 
% (2) \textbf{The usage logs$\footnote{After removing duplicates, there were about 90,000 user data.}$} of our implemented LLMs from six months.
Finally, we include nine tasks, classified into six categories:
% From them, we mine complex instruction data with regard to our defined features via length filtering and regular expression matching.

% (1) \textbf{Complex NLP Tasks}: 人们在real-world scenarios中面临许多NLP任务，但与NLP学术数据集有以下不同：一方面，真实场景的指令往往更多变、有诸多限制，而局限于特定任务的学术数据集可以被很简单的指令描述；另一方面，在真实场景中，输入往往有很多噪音、有很长的上下文，而学术数据集往往是清洗后规范的文本。因此，我们选择了以下NLP任务：\textit{长文本摘要}将多轮录音对话转文字并要求LLMs总结说话人的待办事项。\textit{长文本封闭式问答}根据用户的query利用BM25从PDF转成的文本中查询三段最相关的文本，然后要求模型根据参考文本回答query。\textit{长文本关键词抽取}要求模型根据图片ocr的结果以及图片的上下文结果、教学视频语音转的文本生成关键词。\textit{复杂抽取}要求从新闻文本中抽取指定的关系并以指定的格式返回。
% (2) \textbf{Meta-prompt}: 近年来，人们常通过设计prompt利用大模型帮助构建数据集。这些prompt的指令多样细致、输入的话题丰富、常是few-shot样本、对格式有明确要求且几乎不会在小模型的训练样本中出现。因此，我们搜集了包含但不限于量纲理解、指令优化、复杂抽取、Stable Diffusion prompt生成等任务的元指令。
% (3) \textbf{Planning}: 近年来，研究者们通过设计prompt模仿人们的思维过程指导LLMs完成思考并调度工具完成规划。这些prompt的指令往往限制了可调用的接口、对格式有明确的要求、是few-shot样本、有很长的上下文。因此，我们搜集了基于CN-dbpedia、基金知识库和Langchain完成规划任务的prompt。由于小模型的规划能力有限，在此仅对模型根据观察进行思考并单步规划的结果进行考察。
% \textbf{Structured Input}: 
% (4) \textbf{Well-guided Writing}: 已有benchmark考虑写作方面的评估，但是它们通常没有任何约束且都是单轮的，存在以下问题：忽略了真实场景中用户为了实现高效的写作指导所提出一系列具体要求（比如字数、关键词）；忽略了真实场景中用户很难一次满意，会不断提出修改意见；难以评估。因此，我们构建了带有约束的多轮写作指令，以期模拟用户真实使用过程中对LLMs写作细致的引导。
% (5) \textbf{Detailed Brainstorming}: 人类在跟聊天模型交互时最直观感受的任务类型之一就是头脑风暴，然而现有的评估数据集要么指令太简单开放难以评估、要么太刁钻没有区分度。因此，我们从用户日志中搜集带有多种硬约束的头脑风暴指令。
% \begin{itemize}
    % \item 
    
\textbf{Complex NLP Tasks.} Instructions concerning NLP tasks in real-world scenarios are more diverse and detailed~\cite{xu2023wizardlm} and contain noisy and long contexts~\cite{an2023eval} compared to academic datasets.
% differ from academic datasets in the following aspects.
% On one hand, real-scenario instructions tend to be more diverse and detailed, while academic datasets focusing on certain tasks can be described by simple instructions~\cite{xu2023wizardlm}. 
% On the other hand, real-world inputs frequently contain noisy and long contexts, while academic datasets are typically cleaned and standardized texts~\cite{an2023eval}.
% Overall, we select the four tasks commonly found in existing benchmarks, while incorporating instructions with more complex task descriptions or input beyond those benchmarks:
Overall, we choose four tasks commonly found in existing benchmarks~\cite{liang2022holistic}, enhancing them with more complex instructions and inputs beyond traditional benchmarks:
\textit{long text summarization}, 
\textit{long text closed-domain question answering},
\textit{long text keywords extraction},
\textit{complex information extraction}.
The details can be found in the Appendix.
% (1)~\textit{Long Text Summarization.} requires to summarize the key points from multi-turn audio-converted long conversations.
% (2)~\textit{Long Text Closed-domain Question Answering} aims at answering a user query based on relevant text passages retrieved from PDF-converted text using BM25.
% % (3)~\textit{Long Text Keywords Extraction} requires LLMs to generate keywords for images based on the OCR result of the image and its context from textbooks. \hqy{audio + textbook}
% (3) ~\textit{Long Text Keywords Extraction.} requires to generate keywords based on the context and OCR results of images from textbooks, or speech-to-text transcripts from online courses.
% (4)~\textit{Complex Information Extraction.} aims at retrieving specific information about specific objects from the given text and returning them in a predefined format.

\textbf{Meta-prompt.}
Researchers design elaborate prompts to leverage LLMs to construct datasets~\cite{xu2023wizardlm, honovich2022unnatural, qin2023toolllm}, which can be defined as \textit{Meta-prompts}~\cite{honovich2022unnatural}.
These prompts generally have varied instructions, rich input topics, few-shot samples, clear format requirements and are unlikely to appear in the training samples.
% Therefore, we collect meta-prompts from various novel tasks such as dimension perception, instruction polishment, complex extraction, Stable Diffusion prompt generation, and taxonomy construction.
Therefore, we collect prompts crafted by domain experts who focus on various real-world applications of LLMs, such as financial numerical reasoning and educational knowledge graph taxonomy construction, due to their high quality and origin in real-world scenarios.

\textbf{Planning.}
Many studies have designed prompts to mimic human thinking processes, guiding LLMs to perform reasoning and planning~\cite{yao2023react, liu2023agentbench}.
These prompts often impose restrictions on callable functions, have clear format requirements, offer few-shot samples, and provide long contexts.
Therefore, we collect prompts that require LLMs to complete planning tasks based on CN-DBpedia~\cite{xu2017cn}, fund knowledge base, and those from Langchain$\footnote{https://www.langchain.com/}$.
Since smaller LLMs have limited planning capabilities~\cite{liu2023agentbench}, we solely evaluate the models' ability to perform single-step planning.

% \item 
\textbf{Structured Input.} 
Structured text is a common and crucial type of user input, due to its well-organized and easily interpretable format. 
% Existing evaluation datasets involving structured text either focus on natural language inputs or have limited coverage of structured input types, mainly focusing on Python and C++.
Therefore, we include instructions with:
(1) Six structured data types, namely Markdown, LaTeX, SQL, Tree, Python, JSON.
(2) Two distinct tasks 
% \textit{PathCompose} generates the path to access an element or relationships between elements in the structured data.
% % \textit{PathWalk.} extracts content from the structured data at a specified depth.
% \textit{TextRetrieval} retrieves content from the structured data that satisfies specified requirements.
for their \textit{complexity} and \textit{representativeness}:
\textit{Path Compose} directly evaluates the model's understanding of complex nested data structures, while \textit{TextRetrieval} is a common application to extract content meeting specific requirements.
(3) Two levels of difficulty, which are categorized based on the length and depth of the structured input.

% 主打多轮修改
\textbf{Well-guided Writing.}
% Although existing benchmarks consider the writing-related evaluation, they fail to cover complex features~\cite{zhou2023lima} and are limited to single-turn interactions.
% This results in the following limitations:
Existing benchmarks~\cite{chia2023instructeval} considering writing ability mainly have the following limitations:
(1) They overlook the specific needs users have in real-world scenarios when seeking efficient writing guidance, such as word count, key information, or included hashtags.
(2) They fail to consider the iterative nature of user satisfaction, as users may continually provide modification feedback.
(3) They are difficult to automatically evaluate.
To address these limitations, we collect various single-turn complex instructions covering various complex features and multi-turn instructions that reflect realistic revision needs.

%【TODO】也许也要强调多轮？
\textbf{Detailed Brainstorming.}
% Brainstorming is a typical type of instruction that humans may query with LLMs, consequently yielding an intuitive impression for the chat models.
Brainstorming yields an intuitive impression for the chat models.
However, existing evaluation datasets either have overly simple and open instructions that are difficult to evaluate~\cite{li2023camel}, or they are excessively tricky with limited discrimination$\footnote{https://github.com/zhenbench/z-bench}$.
In our benchmark, we collect single-turn brainstorming data with detailed requirements and multi-turn brainstorming data that simulate realistic user interactions.

\subsubsection{Data Evolution}
% 一些数据增强的方法
% 1. 改任务描述的位置 2. 改任务描述的表述 3. 改prompt格式 4. 改输入文本的格式 5. 增加约束 6. 改成多轮 7. 改few-shot样本数量

% TODO：怎么标数据得写写
% 基于以上selected tasks搜集的复杂指令数据有以下局限性：（1）对于从真实项目中搜集的某类任务的复杂指令，human-curated的任务描述虽然复杂但是雷同。（2）对于从用户使用日志中搜集的数据，有很多没被有效利用的简单指令。因此，我们evolve data for以下目标（1）多样化搜集的复杂指令（2）复杂化简单指令扩大数据规模，从而实现更鲁棒、可信的评估。
The collected complex instructions have two limitations:
(1) For those collected from real-world projects, the human-elaborated task descriptions are complex but alike.
(2) For those collected from usage logs, many simple instructions are not effectively utilized.
% Hence, we evolve our data for the following objectives:
% (1) diversifying the collected complex instructions, 
% (2) complicating the simple instructions to increase the data scale,
% thus achieving a more robust and reliable evaluation.
% 我们从两个视角evolve data: 通过In-breadth Evolution基于现有数据扩充Task Description形式以及内容的多样性，通过In-depth Evolution将已有的简单数据复杂化。基于这两个视角，我们设计了五个维度。For In-breadth evolution, we consider (1) Task Description Relocation: (2) Task Description Paraphrasing: (3) Task Description Imitation: . For In-depth evolution, we consider (1) Constraints Addition: (2) Multi-round Interaction：。根据不同任务的数据特征，我们组合这五个维度设计了三个prompt，分别用于复杂化不同任务的数据（见附录）。并且我们用GPT-3.5-turbo将所有中文数据翻译成英文。
Hence, we introduce two perspectives to evolve data, thereby achieving a more robust and reliable evaluation.
\textbf{In-breadth Evolution} aims to diversify the collected complex instructions (including three methods \textit{task description relocation}, \textit{task description paraphrasing} and \textit{task emulation}).
\textbf{In-depth Evolution} aims to complicate the simple instructions to increase the data scale (including two methods \textit{constraints addition}, \textit{multi-round interaction}).
The motivation and prompts for each method are detailed in the Appendix.

\subsection{Evaluation System}
\subsubsection{Criteria}
% 根据真实场景的需求，我们定义以下硬约束：
% (1) Count limit: the exact or maximum number of words, sentences, or samples allowed in the response，防止答案太过简略或冗余
% (2)* Mandatory words: specific words that must be included in the response, 例如预定义的函数名、答案的主要对象等
% (2) Mandatory words: 任务描述中对答案内容的硬性限制，例如预定义的函数、答案的主要对象、必须包含的关键词等。
% (3) Answer format: the expected structure or format of the answer，例如可解析的json格式、few-shot样本的指定格式等
% (4) Answer type: the expected type of answer，例如在抽取任务中限制抽取的关系为时间、人物、车站名等，在头脑风暴中限制答案的内容范围。感觉还是可以归为Mandatory words
% (5) Content-based query: the requirement that the response should be based on the content of the query itself rather than external sources，which要求模型不能有幻觉，并且能识别已经。

% \input{tables/030labeling}

% 【TODO】画个表格，把hard-contraints高亮出来
% According to the real-world scenario requirements, we define the following hard constraints, as shown in Fig.~\ref{}:
% 根据以上复杂指令的划分维度，我们在评估的过程中考虑以下限制：
We define the following criteria that should be assessed as they can encompass common errors made by models.
(1) \textbf{Count limit}: the number of words, sentences, or samples allowed in the response.
(2) \textbf{Answer format}: the expected structure or format of the response, such as a parsable JSON format, or a specified format for few-shot samples.
(3) \textbf{Task-prescribed phrases}: semantic constraints on the response that are stipulated in the task description, such as predefined functions, primary subjects, or key elements.
(4) \textbf{Input-dependent query}: the query should be answered faithfully according to the given input texts.

Although \textit{Task-prescribed phrases} and \textit{Input-dependent query} both impose content-related constraints on the response, they differ in the information they rely on.
The former centers on constraints explicitly stated by the user in the task description, while the latter focuses on constraints implicitly derived from the content of the input text.
% 这的逻辑是：instruction包含instruction + input。complex包含instruction里本身的限制，也包含input本身的难度。

% (3) \textbf{Mandatory words}: specific words that must be included in the response, such as predefined function names, main objects, or detailed requirements of the answer.

\subsubsection{Evaluation Metrics}
% 针对我们提出的四类硬性约束，我们分别提出了不同视角、不同难度的自动化评估指标，以期构建全面、实用、客观的复杂指令评估系统。
We propose automated evaluation metrics for designed criteria, considering various perspectives and difficulty levels.
% These metrics allow us to evaluate LLMs’ ability to follow complex instructions in a comprehensive, practical, and objective way.
% 对于每一个样本s，它的每一条约束$C=f_n(n, l)$组成，其中$f$是由criterion $n$决定的打分函数，$l$为打分时需要参考的标准，由人工给定。
Each sample $s_i = \{I_i, a_i, h_i\}$ consists of instruction $I_i$, a model answer $a_i$ and given histories$\footnote{To ensure a fair comparison between models, all the model answers in the histories for each sample are the same and provided by GPT-3.5-turbo.}$ $h_i = \{(I_0, a'_0), ..., (I_{i-1}, a'_{i-1})\}$.
Here, $i$ denotes the round number within multi-turn dialogues.
For each sample $s$, its score for each criteria comprises multiple sub-scores $\mathcal{C} = \{c_1, c_2, ..., c_i\}$. 
Each sub-score $c_i = f_{x}(l, a_i, h_i)$ is determined by scoring function $f_n$ based on the criterion $x$, and a limit $l$ manually annotated by humans.
The limit $l$ can be an integer, a list of keywords, or a referenced string$\footnote{The annotation process is detailed in the Appendix.}$.

\input{tables/030statistics_all}
\textbf{Count Limit.}
% 对于单轮指令，我们主要考虑了词数限制、句子数限制、样本数限制。对于多轮指令，我们考虑紧接着的对话是否根据要求进行修改，要求针对句子长短、篇幅长度。
We mainly consider four sub-scores: \textit{word count score}, \textit{sentence count score}, and \textit{sample count score}, \textit{revise score}.
% For multi-turn instructions, we consider whether the subsequent dialogue follows the modification requirements, which are related to \textit{text length}.
% 怎么分词、怎么切分句子、怎么识别样本数、怎么量化篇幅（可以参考一下qxp组的longeval）
% 这里的指标最好能被量化，比如限制5句的话，模型回答10句的分应该比6句的分要低
% \hqy{... (Details)}
For \textit{word count score}$\footnote{Since models can hardly understand the exact word count due to different tokenizers, the exact word count is meaningless.}$
, the criteria can be \textit{word-max} and \textit{word-min}.
For the scoring function $f_{\text{word-max}}$, the more word count exceeds the threshold limit $l_c$, the lower the score will be, thus $f_{\text{word-max}}$ is defined as follows:

\begin{small}
\[f_{\text{word-max}}(a_i, l_c)= 
\begin{cases}
    1 & n(a_i) \leqslant l_c \\
    1 - \frac{|n(a_i) - l|}{n(a_i)} & n(a_i) > l_c
\end{cases}
\]
\end{small}

% 我们只统计不计算标点符号的有效词数

Here, $n(a_i)$ is the valid word count of answer $a_i$ excluding punctuation marks.
$f_{\text{word-min}}$ is defined as follows:

\begin{small}
\[f_{\text{word-min}}(a_i, l_c) = 
\begin{cases}
    1 & n(a_i) \geqslant  l_c \\
    \frac{n(a_i)}{l} & n(a_i) < l_c
\end{cases}
\]
\end{small}

% word_count的得分是word-max和word-min得分的平均值。
% For each sample, the \textit{word count score} is the average of its scores for all criteria (\textit{word-max} and \textit{word-min}).
% scoring functions for sentence count consist of sentence-max, sentence-min, sentence-exact. scoring functions for revise consist of longer and shorter. $f_{\text{revise-longer}} = 1$ when $n(a_i) > n(a_{i-1})$ else  $f_{\text{revise-longer}} = 0$.
Likewise, the scoring functions for \textit{sentence count} encompass $f_{\text{sentence-max}}$, $f_{\text{sentence-min}}$, $f_{\text{sentence-exact}}$.
The scoring function for \textit{sample count} $f_{\text{sample-exact}}$ is implemented using regex matching.
The limit $l_c$ for revise score $f_{\text{revise}}$ can be the string \textit{longer} or \textit{shorter}.
Speicifically, the function $f_{\text{revise}}(a_i, \textit{longer})$ equals 1 if $n(a_i) > n(a_{i-1})$, otherwise, it equals 0.
% 最终，对于一个样本，count limit这一硬约束的整体得分为所有
For each sample, the final \textit{Count Limit} score $S_c$ is the average of all the sub-scores.

% 对于一个样本$s$，

\textbf{Answer Format.}
% 在指令中预定义答案的格式一般是为了自动解析答案。因此，针对答案格式，应该有不同难度的评估。
% 对于每个样本，如果模型生成的结果能够直接通过python的eval()函数或者库（例如xml库）解析，应该给予很高的分数。但是，虽然模型生成的结果不能直接解析，但是模型学到了一定的格式，也证明模型具备遵循复杂指令的能力。因此，对于每条样本，我们从预定义的格式提取了关键词，模型答案覆盖的关键词比例即为最终分数。
% The predefined answer format in the instructions serves to facilitate the automated parsing of the answer~\cite{yao2023react,liu2023agentbench}.
% Hence, we design our evaluation metrics from different levels of difficulty.
% 因此，我们主要按难度划分，考虑了resolve和keywords两个子分数。
This metric has two sub-scores: \textit{parseability} and \textit{keywords}. 
% 首先，如果模型结果可以按照指定的格式解析，则分数为1，否则为0。
First, if the model output can be parsed in the prescribed format, such as JSON, $f_{\text{parseability}}(a_i, \textit{json})$ equals 1; otherwise, it equals 0.
However, even in cases where the model output cannot be directly parsed, its ability to learn certain patterns still demonstrates its capacity to follow complex instructions. 
% 因此，对于每个样本，我们从目标格式中提取关键词列表$l = {w_1, w_2, ..., w_i}$。最终答案对关键词列表的覆盖率。
Consequently, for each sample, we first extract keywords list $l_f = \{w_1, w_2, ..., w_i\}$ from pre-defined formats, which we define as \textit{Scoring Keywords}.
Then, the sub-score $f_{\text{keywords}}(a_i, l_f)$ is defined as follows: 

\begin{small}
\[f_{\text{keywords}}(a_i, l_f) = \frac{N(a_i, l_f)}{|l_f|},
\]
\end{small}
where $N$ denotes the number of scoring keywords covered by the model output $a_i$.
Finally, the overall score for answer format $S_f$ is the average of $f_{\text{parseability}}$ and $f_{\text{keywords}}$.

% 没讲清楚呵呵
% 因为有输入文本的限制，答案的好坏程度可由和正确答案重合的关键信息衡量。类似于老师批阅卷子时，学生回答与答案的关键词重叠越多分数越高，known as 得分点。因此，我们用大模型得到每个Input-dependent query的答案的得分点，模型覆盖的得分点比例即为最终的分数。
\textbf{Input-dependent Query.}
The key phrases of the correct answer stem from the input text.
The more scoring keywords included in a response, the higher the quality of the response.
Hence, for each sample, the subscore $f_{\text{keywords}}(a_i, l)$ is also applied here, where the \textit{Scoring keywords} $l_q$ are extracted from \textit{input text}.
% 特别地，我们发现有些模型在没理解指令时会重复Input-dependent。为了防止这种undesirable copy，我们引入了COPY-BLEU的惩罚项。回答与input-text越相似，则分越低。
Moreover, certain models tend to repeat input text when they fail to understand the instructions, especially when the input text is long and noisy or during the multi-turn dialogue.
To prevent this undesirable copying behavior, we introduce a penalty term known as COPY-BLEU~\cite{chen2022controllable}, which decreases as the response exhibits greater similarity to the input text.
The final score $S_q$ for the Input-dependent query is defined as follows:

% \begin{small}
% \[ S_i = (1 - f_{\text{BLEU}}(a_i, w_i))f_{\text{keywords}}(a_i, l_q),
% \]
% \end{small}
% where $w_i$ is the input text of sample $s_i$. 

% The final score $S_q$ for the Input-dependent query is defined as follows:

\begin{small}
\[ S_q = (1 - f_{\text{BLEU}}(a_i, t_i))f_{\text{keywords}}(a_i, l_q),
\]
\end{small}
where $t_i$ is the input text of sample $s_i$.

\textbf{Task-prescribed Phrases.}
% 对于Task-prescribed phrases而言，任务描述中的强制限制是必须满足的条件，例如抽取任务中限制的对象、关系，写作任务中强制要求需要包含的关键Phrases。答案覆盖的关键Phrases越多，证明模型follow复杂指令的能力越强。
% 为了实现这个目标，对于每个样本，我们人工标注了任务描述中的强制限制短语，作为Scoring keywords。接着，对于每个样本，模型覆盖的Scoring keywords比例为最终的得分。
The mandatory phrases specified in the task description are essential conditions that must be fulfilled.
The more mandatory phrases covered in the answers, the better the model follows complex instructions.
Hence, the subscore $f_{\text{keywords}}(a_i, l_t)$ is applied where $l_t$ is the scoring keywords extracted from the task description.

\subsection{Evaluation of the Benchmark}
Each sample is labeled by three annotators based on our four criteria.
Specifically, we retain samples only when at least two annotators agree on the criteria \textit{Count Limit} and \textit{Output Format Parseability}.
For criteria involving \textit{Keywords Coverage}, we only keep keywords with a consensus from at least two annotators.

\subsection{Statistics of the Benchmark}
% 样本数量、平均Token数量

% 我们将\method的统计信息展示在表\ref{}。不同的任务侧重于不同的constraints。整体而言，我们的数据覆盖了四个constraints。
%The statistics of \method is presented in Tab. \ref{tab:030statistics}. 
% 根据criteria主要分布在task description还是input text，我们的数据集可以分为complex instruction和complex input两类。
%First, our dataset can be classified into two categories, \textit{Complex task description} and \textit{Complex input}, based on the criteria primarily found in the task description or the input text.
%Also, different tasks emphasize different criteria.
%Overall, our dataset effectively covers four criteria.
% 我们对比了\method和其他Benchmark的信息, as shown in Tab. x. 首先，我们的Task Description普遍更长。其次，我们涵盖的任务是Open-ended的，更符合实际应用。最后，我们的评估是客观且细粒度的。
%We also compare existing benchmarks in Tab.~\ref{tab:030statistics_all}. 
%First, our benchmark is the first to systematically investigate LLMs' ability to follow complex instructions.
%The instructions in our benchmark are generally more complex and thus longer.
%Second, the tasks we cover are open-ended, making them more aligned with real-world applications. 
%Moreover, our evaluation is objective and fine-grained.

Tab. \ref{tab:030statistics} presents the statistics$\footnote{Chinese word are counted via https://github.com/fxsjy/jieba. English words are counted via https://www.nltk.org/.}$ of \method. Our dataset has two categories depending on whether the criteria are mainly in the task description or the input text. 
Different tasks also have different emphases on the criteria, and
our dataset covers the four criteria effectively. Tab.~\ref{tab:030statistics_all} compares our benchmark with existing ones. Our benchmark is the first to systematically test LLMs’ ability to follow complex instructions, which are generally longer and more complex than other benchmarks. The tasks we cover are open-ended, which are more realistic and practical. 
Our evaluation is also more objective and fine-grained.

%% file: tables/030statistics.tex
\begin{table*}[t]
  \centering
  \small
  \resizebox{0.75\textwidth}{!}{
\begin{tabular}{cccccccccc}
\toprule
\textbf{Category}                                                                        & \textbf{Tasks}      & \textbf{\#Samples} & \textbf{\#Format} & \textbf{\#Task} & \textbf{\#Input} & \textbf{\#Count} & \textbf{Avg TD Len.} & \textbf{Avg IP Len.} & \textbf{Avg Ins Len.} \\
\midrule
\multirow{5}{*}{\textbf{\begin{tabular}[c]{@{}c@{}}Complex \\ Task \\ Description\end{tabular}}} 
& \textbf{Extraction}  & 49         & 49       & 35     & 49      & N/A      & 125         & 169         & 295          \\
                                                                                         & \textbf{Planning}  & 52         & 52       & 46     & 48      & N/A       & 1070        & 534         & 1606         \\
                                                                                         & \textbf{Meta.}     & 20         & 20       & 15     & 6       & 2       &    765         & 166         & 933          \\
                                                                                         & \textbf{BS(S)}      & 20         & 20       & 20     & 1       & 15      & 70          & N/A           & 70           \\
                                                                                         & \textbf{Writing(S)}  & 23         & 2        & 23     & 2       & 12      & 82          & 25          & 107          \\
                                                 \midrule
\multirow{6}{*}{\textbf{\begin{tabular}[c]{@{}c@{}}Complex \\ Input\end{tabular}}}       & \textbf{Keywords}   & 15         & 15       & 15     & 15      & N/A       & 546         & 943         & 1579         \\
                                                                                         & \textbf{QA}       & 89         & N/A        & N/A      & 89      & N/A       & 25          & 881         & 814          \\
                                                                                         & \textbf{Sum.}    & 108        & N/A        & N/A      & 108     & N/A       & 45          & 514         & 562          \\
                                                                                         & \textbf{Struture}  & 38         & 6        & N/A      & 38      & N/A       & 29          & 1360        & 1390         \\
                                                                                         & \textbf{BS(M)}     & 52         & 50       & 50     & 10      & 36      & 31          & 559         & 31           \\
                                                                                         & \textbf{Writing(M)} & 57         & 3        & 35     & 48      & 43      & 30          & 656         & 51           \\
                                                                                         \midrule
\multicolumn{2}{c}{\textbf{Overall}}                                                                           & 523        & 217      & 239    & 414     & 108     & 256         & 528         & 676                  \\
\bottomrule
\end{tabular}
}

  \caption{The statistics of our benchmark. For each task, \#Format, \#Task, \#Input, \#Count denote the number of samples covering the criteria \textit{Answer format}, \textit{Task-prescribed phrases}, \textit{Input-dependent query}, and \textit{Count limit} respectively. 
  Avg TD/IP/Ins Len. denote the average word number of \textit{task description}, \textit{input text} and \textit{instruction}.
  Meta., BS, SUM. denote the Meta-prompt, Brainstorming, Summarization task respectively. 
  (S) and (M) represent single-round and multi-round.
  N/A denotes that such tasks do not involve corresponding evaluation criteria.
  }
  \label{tab:030statistics}
\end{table*}

%% file: tables/030statistics_all.tex
\begin{table}[t]
  \centering
  \scriptsize
  \resizebox{0.48\textwidth}{!}{
\begin{tabular}{ccccccc}
\toprule
\textbf{Benchmark}                                             & \textbf{Focus}                                                                     & \textbf{\begin{tabular}[c]{@{}c@{}}Avg Ins \\ Len.\end{tabular}} & \textbf{Format}    & \textbf{Evaluation}                                                     & \textbf{Objective} \\
\midrule
% MMLU                                                                      & Knowledge                                                                          & 53                                                                  & C                  & Acc.                                                                    & T                  \\
\textbf{C-Eval}               & Knowledge                                                                          & 110                                                                  & C                  & ACC                                                                    & T                  \\

\midrule

\textbf{AGIEval}                                                            & Knowledge                                                                          & 184                                                                 & C                  & EM/F1                                                                   & T                  \\

\midrule
\multirow{2}{*}{\textbf{Kola}}                                       & \multirow{2}{*}{Knowledge}                                                         & \multirow{2}{*}{310}                                                & C                  & \begin{tabular}[c]{@{}c@{}}EM/F1\\ /ACC\end{tabular}                    & T                  \\

\cmidrule(lr){4-6}
                                                           &                     &                                                                                    &                                                                O                  & BLEU/Rouge                                                              & T                  \\
% \multirow{2}{*}{L-Eval}                                    & \multirow{2}{*}{\begin{tabular}[c]{@{}c@{}}Long-Text\\ Understanding\end{tabular}} & 5872                                                                & C                  & \begin{tabular}[c]{@{}c@{}}EM/F1\\ /Rouge\end{tabular}                  & T                  \\
%                                                            &                     &                                                                                    13075                                                               & O                  & Preference             & F                  \\

\midrule
\begin{tabular}[c]{@{}c@{}}\textbf{WizardLM} \\\textbf{Testset}\end{tabular}               & \begin{tabular}[c]{@{}c@{}}Complex\\ Instruction\end{tabular}          & 62                                                                  & O                  & Preference             & F                  \\

\midrule
\multirow{2}{*}{\textbf{ToolBench}}                                 & \multirow{2}{*}{Planning}                                                          & \multirow{2}{*}{N/A}                                                & \multirow{2}{*}{O} & Pass Rate                                                               & T                  \\
\cmidrule(lr){5-6}
                                                           &                     &                                                                                    &                                                                     &                  Preference                                                              & F                  \\

                                                           \midrule
\textbf{AgentBench}                                                              & \begin{tabular}[c]{@{}c@{}}Desicion\\ Making\end{tabular}                          & N/A                                                                 & O                  & Pass Rate                                                               & T                  \\

\midrule
\textbf{HumanEval}                                                             & Programming                                                                        & N/A                                                                 & O                  & Pass Rate                                                               & T                  \\
\midrule
\textbf{\method}                                                               & \begin{tabular}[c]{@{}c@{}}Complex \\ Instruction\end{tabular}        &    676                                                                 & O                  & \begin{tabular}[c]{@{}c@{}}Four \\ Fine-grained \\ Metrics\end{tabular} & T      \\           
\bottomrule
\end{tabular}
}
  \caption{Statistics of existing benchmarks.
  %: C-Eval~\cite{huang2023c}, AGIEval~\cite{zhong2023agieval}, Kola~\cite{yu2023kola}, WizardLM~\cite{xu2023wizardlm}, ToolBench~\cite{qin2023toolllm}, AgentBench~\cite{liu2023agentbench}, HumanEval~\cite{chen2021evaluating}. 
  \textit{Avg Ins} denotes the average word numbers in instructions. 
  C and O denotes the Close-ended and Open-ended respectively.
  \textit{Preference} refers to evaluation via GPT4.
  \textit{Objective} represents whether the evaluation metrics are objective (T) or subjective (F).}
  \label{tab:030statistics_all}
\end{table}

%% file: 040experiment.tex
\input{tables/040overall}

\input{tables/040constraints}

\input{tables/040discrimination}
\input{tables/040visual}
\section{Experiment}
\paragraph*{Evaluated Models}
% 大模型
% 中文小模型
% 英文小模型

% 我们评估了xx个在其他榜单表现出色的模型，varying from他们的模型大小、支持的上下文长度、instruction tuning data等，as shown in Tab. 2.
% 我们将模型划分为三组：English or Chinese-oriented Models是指预训练语料中只有minor fraction的中文数据以及大量中文数据的模型。Chinese-oriented Models (From Scratch, FS) 是指基座模型是基于中文语料从头训练的大模型。Chinese-oriented Models (Continue Pretraining, CP)是指接着English-oriented Models (通常是Llama)继续在中文语料上预训练的大模型。
We evaluate a total of \modelNum models that demonstrated exceptional performance on other benchmarks~\cite{huang2023c, dubois2023alpacafarm, zhong2023agieval}, ranging from their model size, supported context length, and instruction tuning data size, as illustrated in Appendix.
These models are categorized into three groups: Chinese-oriented Models~\textit{(From Scratch, FS)}, Chinese-oriented Models~\textit{(Continue Pretraining, CP)}, and English-oriented Models. 
The distinction between English and Chinese-oriented Models lies in the composition of their pretraining corpus, whereby the former possesses a small portion and the latter possesses a substantial volume of Chinese data.
Chinese-oriented Models~\textit{(FS)} are trained entirely from scratch using Chinese corpora. 
Chinese-oriented Models~\textit{(CP)} continue pretraining on Chinese corpora utilizing an English-oriented base model.

%\subsection{Results}
\paragraph*{Task-categorized Performance}
The performance of the models on different tasks is shown in Tab.~\ref{tab:040overall}.
% 我们首先对比了模型在不同任务上的表现，as shown in Tab. xx. 任务分为Complex Instruction以及Complex Input。

\textit{General Comparisons.}
% 整体而言，chatglm1表现最好，其次是longchat7b 32k，以及chatglm2。对比每组模型的最好模型，Chinese-oriented Models (From Scratch)最佳，English-oriented Models其次，随后是Chinese-oriented Models (cc)。整体而言，Benchmark对模型的区分度较大（discriminative），10B附近模型的表现会从10到70之间。
%First, among the models assessed, Llama2-OpenBuddy emerged as the top performer, closely followed by Vicuna33b and ChatGLM1.
Among the models assessed, OpenChat-V3.2 was the best, followed by Vicuna-V1.5-13B and ChatGLM.
These models had different parameter sizes (13B, 6B), showing that small-scale LLMs can follow complex instructions as well as larger ones. 
The Chinese-oriented (FS) group and the English-oriented group perform equally well and better than the Chinese-oriented (CC) group, proving that complex instruction comprehension is not language-dependent. 
%All three models possess different parameter sizes (13B, 33B, 6B), which indicates that small-scale LLMs can follow complex instructions similarly to their significantly larger counterparts.
% 10B规模附近的模型表现得比33b甚至70b的模型显著更好，这证明小模型
%Additionally, when comparing the overall performance of each group, the Chinese-oriented (FS) group and the English-oriented group exhibit similarly outstanding performance, surpassing that of the Chinese-oriented (CC) group.
%This proves that the comprehension of complex instructions is not limited to the language of the training data.
% 这证明复杂指令的理解并不局限于training data的语言。
% 最后，基于同样的基座模型以及词表的情况下，模型的表现相差很大，例如用llama2-7b作为基座模型、词表以及supported context length，llama2-7b-chat, llama2-linksoul, llama2-flagalpha的表现分布于0.409到0.645之间，这证明复杂指令理解能力与instruction tuning阶段高度相关。
Moreover, under the same base model, vocabulary, and supported context length (e.g. Llama2-7B), the performance of the models varies greatly (e.g. Llama2-chat-7B, Llama2-LinkSoul, and Llama2-FlagAlpha).
This demonstrates a strong correlation between the ability to comprehend complex instructions and the instruction tuning phase.
% 整体而言，现有的开源的中小规模的表现仍然差于闭源的大模型。
Overall, the current open-source small to medium-scale models exhibit a significant performance gap compared to close-source large-scale models (GPT-3.5-turbo, GPT4).

\textit{Complex Task Description.}
% 首先，对于复杂任务描述的中文数据集而言，openchat13b表现得最好，其次是vicuna33b，它们都属于English-oriented Models。这证明对复杂指令的理解在不同语言间有一定可迁移性。
% 其次，对于同一系列模型而言，更大的模型规模普遍带来了提升，例如vicuna(7b, 13b, 33b), vicuna-16k(7b, 13b), llama2-chat(13b, 70b), alpaca(13b, 33b), baize(7b, 13b).
% 接着，每组表现最好的模型（chatglm, llama2-linly, vicuna33b）的supported context length均是2048，这证明更长的上下文文本建模能力和复杂任务描述的理解能力没有显著关系。
Among the data with complex task descriptions, 
% first, OpenChat-v3.2 performs the best, followed by ChatGLM1 and Vicuna33b.
first, four of the top 5 models belong to the English-oriented Models, which demonstrate that the ability to understand complex task descriptions can transfer across different languages. 
Next, within the same series of models, larger model sizes do not always lead to improvements.
% This can be observed in models such as Vicuna (7b, 13b, 33b), Vicuna16k (7b, 13b), Llama2-chat (7b, 13b), Alpaca (13b, 33b).
% Furthermore, the best-performing models in each group (Llama2-OpenBuddy, ChatGLM1, OpenChat-v3.2) have a supported context length of less than 4096.
Furthermore, the best-performing models in each group have a supported context length of less than 4096, suggesting that the supported text context length does not significantly impact the ability to comprehend complex task descriptions.
% Lastly, under the same base model, llama7b, and its vocabulary, there is a notable difference in performance between llama2-7b-chat, vicuna7b-v1.3, llama2-linksoul, and llama2-flagalpha. 
% This demonstrates a strong correlation between the ability to understand complex task descriptions and the instruction tuning stage.

\textit{Complex Input Text.}
% 首先，对于复杂输入的中文数据集而言，排名前5的模型是chatglm1, longchat7b-32k, chatglm2, chatglm2-32k, llama2-linksoul and llama2-openbuddy，其中四个模型都属于Chinese-oriented model，这意味着训练阶段见更多的中文语料规模会帮助模型理解长且有噪音的中文文本。
% 其次，对于English-oriented Models而言，更大的模型规模普遍损害了模型的表现，例如llama2-chat (13b, 70b)，vicuna-v1.5 (7b, 13b)。而对于Chinese-oriented Models而言，更大的模型规模提升了模型的表现，例如Baize(7b, 13b), alpaca(7b, 13b, 33b)
% 最后，Longer supported context length和long, noisy的输入文本理解之间没有显著的关系，chatglm2(4k, 32k)持平, vicuna7b(2k, 16k), vicuna13b(2k, 16k)下降，longchat7b(16k, 32k)上升。
For the data with complex input text, first, seven of the top 10 models belong to Chinese-oriented models, which implies that more Chinese training data assists the models in comprehending long and noisy Chinese texts.
% Next, for English-oriented models, larger model scales generally result in performance drops, such as Llama2-chat (13b, 70b) and vicuna-v1.5 (7b, 13b).
Next, within the same model series, larger scales generally improve performance, while longer supported context length can result in performance drops in many cases.
% , suggesting that an extended context length does not inherently facilitate the model to comprehend noisy input text.
% The model chatglm2 (4k, 32k) remains consistent, while vicuna7b (2k, 16k) and vicuna13b (2k, 16k) experience a decline in performance, and longchat7b (16k, 32k) shows an improvement.

\paragraph*{Criteria-categorized Performance}

As shown in Tab.~\ref{tab:040constraints}, regarding \textit{Answer format}, the English-oriented Models significantly perform better than Chinese-oriented Models.
This demonstrates the English-oriented Models' ability to follow few-shot examples and generate code, as well as partially explains why their complex instruction-following ability can transfer across languages. 
Next, for \textit{Task-prescribed phrases}, two of the top-3 models are Chinese-oriented Models, suggesting that Chinese data helps the models understand Chinese semantic restrictions.
Finally, the performance differences between models for \textit{Count limit} criteria are not big compared to other criteria, which shows that the models have similar comprehension of numerical concepts.

% 对比现有的benchmark的排名

\paragraph*{Comparisons between Benchmarks}
% 我们对比了有代表性的模型在不同主流benchmark上的表现效果，as shown in Fig.~\ref{}。首先，在ceval, CMMLU，GAOKAO这几个重点关注知识的Chinese-oriented benchmark上，许多小模型已经达到和GPT-3.5-turbo相似甚至更佳的水平。其次，在复杂推理（BBH，GSM8k）、编程能力（HumanEval）等challenging benchmarks上，小模型彼此之间缺乏区分度。整体而言，在我们的Benchmark上，小模型的排名更有区分度。
We present the performance$\footnote{https://opencompass.org.cn/leaderboard-llm.}$ of representative models on mainstream benchmarks in Fig.~\ref{fig:040discrimination}. 
First, on benchmarks focusing on Chinese knowledge (C-eval, CMMLU, and GAOKAO), smaller models achieve similar or even better performance compared to GPT-3.5-turbo. 
Also, on challenging benchmarks like complex reasoning (BBH, GSM8k) and programming ability (HumanEval), there is a lack of distinction between smaller models. 
Overall, our benchmark can exhibit more discriminative results.

\paragraph*{Fine-grained Evaluation}
% 我们将有代表性的模型按照任务和constraints维度的表现效果展示至图~\ref{}。首先，根据图~\ref{}左，可以看到不同模型有自己擅长的任务，chatglm2-32k虽然在复杂的task description上表现不好，但是在复杂中文输入文本的任务上表现非常好。其次，根据图~\ref{}左，可以看到不同模型有自己擅长的constraint，llama2-7b-chat在理解format的能力非常强，但是在理解中文的Task和Input constraints时表现不佳。
%The performance of LLMs grounded on the same base model~\cite{touvron2023llama} regarding different tasks and criteria is depicted in Fig.~\ref{fig:040visual}. 
%First, different models exhibit distinct strengths across different criteria.
%For instance, Llama2-7b-chat demonstrates remarkable ability in understanding format but performs poorly in comprehending Chinese input and semantic constraints.
%Next, different models excel in specific tasks. 
%Despite performing poorly in the complex input text, Llama2-7b-chat exhibits well performance in handling complex task descriptions.

Fig.~\ref{fig:040visual} shows the performance of LLMs based on the same base model for different tasks and criteria. 
Different models have different strengths for different criteria. For example, Llama2-chat-7B is good at understanding format but bad at comprehending Chinese input and semantic constraints. 
Different models also excel in specific tasks. 
Llama2-chat-7B handles complex task descriptions well, but not complex input text.

%% file: tables/040overall.tex
\newcolumntype{b}{>{\columncolor{Blue!10}}r}
\newcolumntype{d}{>{\columncolor{brown!10}}r}
\newcolumntype{q}{>{\columncolor{Green!10}}r}

\setlength\tabcolsep{4pt}
\begin{table*}[t]
  \centering
  \scriptsize
    \resizebox{1\textwidth}{!}{
\begin{tabular}{ccccccdccccccqb}
\toprule
\multirow{2}{*}{\textbf{Model}} & \multicolumn{6}{c}{\textbf{Complex Task Description}}                                                                   & \multicolumn{7}{c}{\textbf{Complex Input}}                                                                                        & \multicolumn{1}{c}{\textbf{All}}  
\\
  \cmidrule(lr){2-7} \cmidrule(lr){8-14}\cmidrule(lr){15-15}
                                & \textbf{Extraction} & \textbf{Planning} & \textbf{Meta.} & \textbf{Writing(S)} & \textbf{BS(S)} & \textbf{Average} & \textbf{Keywords} & \textbf{QA}    & \textbf{Sum.}  & \textbf{Struture} & \textbf{Writing(M)} & \textbf{BS(M)} & \textbf{Average} & \textbf{Average} \\
               
% \multicolumn{15}{c}{\textbf{Chinese-oriented Models (Continue Pretraining)}}                                                                                                                                                                                                                                \\
   \midrule
    \rowcolor[gray]{0.95}\multicolumn{15}{c}{\textit{Chinese-oriented Models (Continue Pretraining)}} \\

\textbf{Baize-V2-7B}           & 0.203          & 0.266          & 0.300          & 0.504          & 0.245          & 0.304          & 0.056          & 0.121          & 0.045          & 0.593          & 0.381          & 0.558          & 0.292          & 0.298          \\
\textbf{Llama2-FlagAlpha}      & 0.205          & 0.095          & 0.129          & 0.262          & 0.547          & 0.248          & 0.150          & 0.423          & 0.297          & 0.354          & 0.406          & 0.591          & 0.370          & 0.309          \\
\textbf{Baize-V2-13B}          & 0.214          & 0.334          & 0.342          & 0.272          & 0.536          & 0.340          & 0.070          & 0.143          & 0.019          & 0.540          & 0.433          & 0.574          & 0.296          & 0.318          \\
\textbf{Chinese-Alpaca-V1-13B} & 0.289          & 0.183          & 0.209          & 0.209          & 0.697          & 0.317          & 0.411          & 0.272          & 0.226          & 0.399          & 0.291          & 0.480          & 0.347          & 0.332          \\
\textbf{Chinese-Alpaca-V1-7B}  & 0.264          & 0.123          & 0.215          & 0.357          & 0.612          & 0.314          & 0.265          & 0.267          & 0.243          & 0.465          & 0.401          & 0.703          & 0.391          & 0.352          \\
\textbf{Llama2-Linly}          & 0.382          & 0.170          & 0.205          & 0.352          & 0.527          & 0.327          & 0.196          & 0.464          & 0.406          & 0.596          & 0.352          & 0.594          & 0.435          & 0.381          \\
\textbf{Chinese-Alpaca-V1-33B} & 0.379          & 0.200          & 0.283          & 0.664          & 0.663          & 0.438          & 0.415          & 0.334          & 0.221          & 0.426          & 0.476          & 0.609          & 0.413          & 0.426          \\
\textbf{BELLE}                 & 0.400          & 0.157          & 0.363          & 0.589          & 0.734          & 0.449          & 0.379          & 0.478          & 0.508          & 0.458          & 0.439          & 0.672          & 0.489          & 0.469          \\
\textbf{CuteGPT}               & 0.482          & 0.529          & 0.460          & 0.534          & 0.739          & 0.549          & 0.294          & 0.506          & 0.459          & 0.653          & 0.626          & 0.804          & 0.557          & 0.553          \\
\textbf{Llama2-LinkSoul}       & 0.521          & 0.326          & 0.431          & 0.652          & 0.769          & 0.540          & 0.615          & \textbf{0.788} & 0.684          & 0.565          & 0.747          & \underline{ 0.909}    & 0.718          & 0.629          \\
\textbf{Llama2-OpenBuddy}      & 0.585          & 0.638          & 0.344          & 0.697          & 0.697          & 0.592          & 0.638          & 0.752          & 0.685          & \textit{0.711} & \textbf{0.812} & 0.892          & \textbf{0.748} & 0.670          \\

   \midrule
    \rowcolor[gray]{0.95}\multicolumn{15}{c}{\textit{Chinese-oriented Models (From Scratch)}} \\                                                                                                                                     
\textbf{BatGPT-sirius}         & 0.011          & 0.044          & 0.094          & 0.352          & 0.233          & 0.147          & 0.046          & 0.394          & 0.054          & 0.294          & 0.135          & 0.321          & 0.207          & 0.177          \\
\textbf{MOSS}                  & 0.493          & 0.310          & 0.461          & 0.634          & 0.644          & 0.508          & 0.473          & 0.396          & 0.500          & 0.521          & 0.696          & 0.658          & 0.541          & 0.525          \\
\textbf{InternLM}              & 0.452          & 0.540          & 0.493          & 0.690          & 0.622          & 0.559          & 0.247          & 0.515          & 0.399          & 0.428          & 0.732          & 0.877          & 0.533          & 0.546          \\
\textbf{ChatGLM2}              & 0.539          & 0.317          & \textit{0.608} & 0.664          & 0.632          & 0.552          & 0.589          & 0.725          & 0.669          & 0.590          & 0.738          & 0.777          & 0.681          & 0.616          \\
\textbf{ChatGLM2-32k}          & 0.526          & 0.399          & 0.572          & 0.699          & 0.690          & 0.577          & 0.653          & 0.686          & 0.571          & 0.427          & 0.758          & 0.876          & 0.662          & 0.620          \\
\textbf{Baichuan-chat}     & 0.473          & 0.373          & 0.471          & \textbf{0.800} & 0.794          & 0.582          & 0.491          & 0.728          & \textit{0.701} & 0.601          & \underline{ 0.776}    & 0.857          & 0.692          & 0.637          \\
\textbf{Qwen}                  & 0.544          & 0.551          & 0.493          & 0.646          & 0.740          & 0.595          & 0.486          & \textit{0.767} & \underline{ 0.705}    & 0.575          & 0.710          & 0.888          & 0.689          & 0.642          \\
\textbf{ChatGLM}               & \textbf{0.649} & 0.522          & \underline{ 0.612}    & 0.700          & 0.808          & \underline{ 0.658}    & 0.532          & 0.742          & 0.672          & 0.573          & 0.735          & 0.870          & 0.687          & \textit{0.673} \\

   \midrule
   \rowcolor[gray]{0.95}\multicolumn{15}{c}{\textit{English-oriented Models}} \\     
\textbf{Llama2-chat-7B}        & 0.495          & 0.326          & 0.500          & 0.358          & 0.465          & 0.429          & 0.157          & 0.135          & 0.060          & 0.708          & 0.541          & 0.447          & 0.341          & 0.385          \\
\textbf{Llama2-chat-70B}       & 0.431          & 0.289          & 0.484          & 0.397          & 0.472          & 0.415          & 0.147          & 0.158          & 0.079          & \underline{ 0.719}    & 0.570          & 0.552          & 0.371          & 0.393          \\
\textbf{Llama2-chat-13B}       & 0.445          & 0.329          & \textbf{0.624} & 0.359          & 0.453          & 0.442          & 0.154          & 0.127          & 0.108          & \textbf{0.753} & 0.569          & 0.458          & 0.361          & 0.402          \\
\textbf{Vicuna-V1.3-7B}        & 0.485          & 0.661          & 0.303          & 0.748          & 0.665          & 0.573          & 0.180          & 0.651          & 0.583          & 0.525          & 0.674          & 0.773          & 0.564          & 0.569          \\
\textbf{WizardLM}              & 0.422          & 0.592          & 0.281          & 0.675          & \underline{ 0.856}    & 0.565          & 0.261          & 0.594          & 0.570          & 0.519          & 0.711          & 0.839          & 0.582          & 0.574          \\
\textbf{LongChat-V1-13B}       & 0.523          & 0.591          & 0.423          & 0.654          & 0.533          & 0.545          & 0.400          & 0.572          & 0.532          & 0.579          & 0.752          & 0.810          & 0.607          & 0.576          \\
\textbf{LongChat-V1.5-7B}      & 0.489          & 0.620          & 0.358          & 0.664          & 0.731          & 0.572          & 0.608          & 0.687          & 0.633          & 0.378          & 0.747          & 0.825          & 0.646          & 0.609          \\
\textbf{LongChat-V1-7B}        & 0.549          & 0.475          & 0.424          & 0.710          & 0.805          & 0.593          & 0.527          & 0.604          & 0.557          & 0.692          & 0.729          & 0.856          & 0.661          & 0.627          \\
\textbf{Vicuna-V1.3-13B}       & 0.521          & 0.625          & 0.474          & 0.743          & \textit{0.840} & 0.641          & 0.346          & 0.672          & 0.582          & 0.613          & 0.651          & 0.869          & 0.622          & 0.631          \\
\textbf{Vicuna-V1.5-7B}        & 0.544          & 0.670          & 0.398          & 0.506          & 0.770          & 0.578          & \underline{ 0.711}    & 0.739          & 0.667          & 0.513          & 0.693          & \textit{0.906} & 0.705          & 0.641          \\
\textbf{Vicuna-V1.3-33B}       & 0.589          & \textit{0.702} & 0.385          & \textit{0.752} & 0.835          & 0.653          & 0.503          & 0.680          & 0.643          & 0.627          & 0.622          & 0.872          & 0.658          & 0.655          \\
\textbf{Vicuna-V1.5-13B}       & \textit{0.601} & \underline{ 0.721}    & 0.425          & 0.744          & 0.794          & \textit{0.657} & \textit{0.682} & 0.765          & \textbf{0.723} & 0.630          & 0.746          & 0.896          & \textit{0.740} & \underline{ 0.699}    \\
\textbf{OpenChat-V3.2}         & \underline{ 0.629}    & \textbf{0.733} & 0.510          & \underline{ 0.754}    & \textbf{0.868} & \textbf{0.699} & \textbf{0.725} & \underline{ 0.771}    & 0.663          & 0.608          & \textit{0.761} & \textbf{0.919} & \underline{ 0.741}    & \textbf{0.720} \\
  \cdashlinelr{1-15}
\textbf{GPT-3.5-turbo}         & 0.709          & 0.805          & 0.632          & 0.879          & 0.854          & 0.776          & 0.765          & 0.795          & 0.832          & 0.697          & 0.879          & 0.908          & 0.813          & 0.794          \\
\textbf{GPT-4}                 & 0.737          & 0.879          & 0.666          & 0.828          & 0.810          & 0.784          & 0.862          & 0.889          & 0.911          & 0.727          & 0.867          & 0.910          & 0.861          & 0.822      \\    
\bottomrule
\end{tabular}
}
  \caption{The performance of models on different tasks. 
  Detailed information of each model is provided in the Appendix. 
  % 加粗、下划线、斜体分别代表排名第一、第二、第三的结果
  The \textbf{bold}, \underline{underlined}, and \textit{italicized} denote the first, second, and third rankings, respectively.
  }
  \label{tab:040overall}
\end{table*}

%% file: tables/040constraints.tex
\newcolumntype{a}{>{\columncolor{BlueGreen!10}}c}
\newcolumntype{b}{>{\columncolor{Blue!10}}r}
\newcolumntype{d}{>{\columncolor{brown!10}}r}
\newcolumntype{q}{>{\columncolor{Green!10}}r}

\begin{table}[!htb]
  \centering
  \scriptsize
  % \resizebox{\textwidth}{!}{
  \renewcommand\arraystretch{0.9}
    \resizebox{0.4\textwidth}{!}{
\begin{tabular}{ccccca}
\toprule
\textbf{Model}      & \textbf{Format} & \textbf{Input} & \textbf{Task}  & \textbf{Count} & \textbf{Average} \\
\midrule
    \rowcolor[gray]{0.95}\multicolumn{6}{c}{\textit{Chinese-oriented Models (Continue Pretraining)}} \\

\textbf{Baize-V2-7B}           & 0.409           & 0.300          & 0.246          & 0.466          & 0.298            \\
\textbf{Llama2-FlagAlpha}      & 0.499           & 0.218          & 0.221          & 0.468          & 0.309            \\
\textbf{Baize-V2-13B}          & 0.530           & 0.247          & 0.302          & 0.444          & 0.318            \\
\textbf{Chinese-Alpaca-V1-13B} & 0.603           & 0.207          & 0.259          & 0.458          & 0.332            \\
\textbf{Chinese-Alpaca-V1-7B}  & 0.663           & 0.224          & 0.256          & 0.512          & 0.352            \\
\textbf{Llama2-Linly}          & 0.411           & 0.347          & 0.374          & 0.490          & 0.381            \\
\textbf{Chinese-Alpaca-V1-33B} & 0.655           & 0.353          & 0.357          & 0.576          & 0.426            \\
\textbf{BELLE}                 & 0.556           & 0.408          & 0.484          & 0.498          & 0.469            \\
\textbf{CuteGPT}               & 0.640           & 0.548          & 0.576          & 0.514          & 0.553            \\
\textbf{Llama2-LinkSoul}       & 0.662           & 0.623          & 0.662          & 0.603          & 0.629            \\
\textbf{Llama2-OpenBuddy}      & 0.734           & 0.627          & 0.704          & 0.638          & 0.670            \\

\midrule
    \rowcolor[gray]{0.95}\multicolumn{6}{c}{\textit{Chinese-oriented Models (From Scratch)}}                                                       \\
\textbf{BatGPT-sirius}         & 0.154           & 0.206          & 0.069          & 0.357          & 0.177            \\
\textbf{MOSS}                  & 0.586           & 0.514          & 0.564          & 0.534          & 0.525            \\
\textbf{InternLM}              & 0.650           & 0.527          & 0.524          & 0.612          & 0.546            \\
\textbf{ChatGLM2}              & 0.620           & 0.605          & 0.691          & 0.568          & 0.616            \\
\textbf{ChatGLM2-32k}          & 0.687           & 0.563          & \textit{0.716} & 0.603          & 0.620            \\
\textbf{Baichuan-chat}     & 0.750           & 0.603          & 0.586          & \textit{0.662} & 0.637            \\
\textbf{Qwen}                  & 0.764           & 0.584          & 0.625          & 0.570          & 0.642            \\
\textbf{ChatGLM}               & 0.715           & \textit{0.628} & \underline{ 0.742}    & 0.571          & \textit{0.673}   \\

\midrule
    \rowcolor[gray]{0.95}\multicolumn{6}{c}{\textit{English-oriented Models}} \\                                                                            
\textbf{Llama2-chat-7B}        & 0.598           & 0.294          & 0.306          & \underline{ 0.686}    & 0.385            \\
\textbf{Llama2-chat-70B}       & 0.631           & 0.318          & 0.265          & \textbf{0.701} & 0.393            \\
\textbf{Llama2-chat-13B}       & 0.640           & 0.342          & 0.280          & 0.674          & 0.402            \\
\textbf{Vicuna-V1.3-7B}        & 0.598           & 0.520          & 0.599          & 0.597          & 0.569            \\
\textbf{WizardLM}              & 0.730           & 0.525          & 0.531          & 0.586          & 0.574            \\
\textbf{LongChat-V1-13B}       & 0.723           & 0.528          & 0.585          & 0.507          & 0.576            \\
\textbf{LongChat-V1.5-7B}      & \textbf{0.791}  & 0.518          & 0.589          & 0.535          & 0.609            \\
\textbf{LongChat-V1-7B}        & \underline{ 0.789}     & 0.574          & 0.615          & 0.609          & 0.627            \\
\textbf{Vicuna-V1.3-13B}       & 0.766           & 0.588          & 0.641          & 0.554          & 0.631            \\
\textbf{Vicuna-V1.5-7B}        & 0.756           & 0.536          & 0.698          & 0.599          & 0.641            \\
\textbf{Vicuna-V1.3-33B}       & 0.770           & 0.609          & 0.668          & 0.575          & 0.655            \\
\textbf{Vicuna-V1.5-13B}       & \textit{0.786}  & \underline{0.656}          & 0.701          & 0.640          & \underline{ 0.699}      \\
\textbf{OpenChat-V3.2}         & 0.766           & \textbf{0.703} & \textbf{0.776} & 0.617          & \textbf{0.720}   \\

  \cdashlinelr{1-6}
\textbf{GPT-3.5-turbo}         & 0.899           & 0.760          & 0.799          & 0.700          & 0.794            \\
\textbf{GPT-4}                 & 0.911           & 0.796          & 0.792          & 0.724          & 0.822          \\
\bottomrule
\end{tabular}
}
  \caption{The performance of models regarding different criteria.  The \textbf{bold} and \underline{underlined}, and \textit{italicized} denote the first, second, and third rankings, respectively.}
  \label{tab:040constraints}
\end{table}

% 还应该加个chatgpt

%% file: tables/040discrimination.tex
\begin{figure}[t] 
    \centering
        \includegraphics[width=0.75\linewidth]{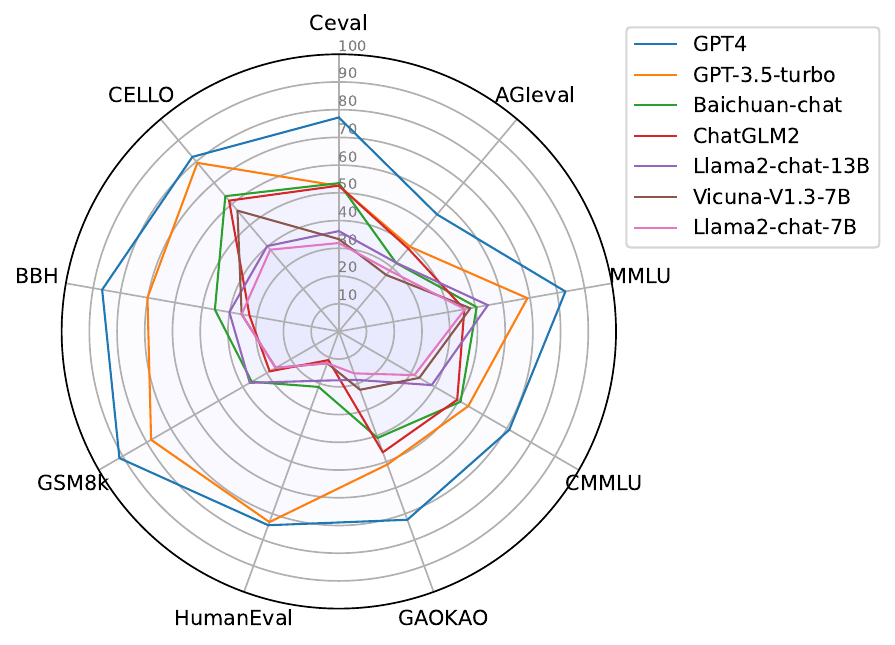} 
    \captionsetup{font={small}} 
    \caption{The performance of models on mainstream benchmarks.}
    \label{fig:040discrimination}
\end{figure}

%% file: tables/040visual.tex
\begin{figure}[t] 
    \centering
        \includegraphics[width=1.0\linewidth]{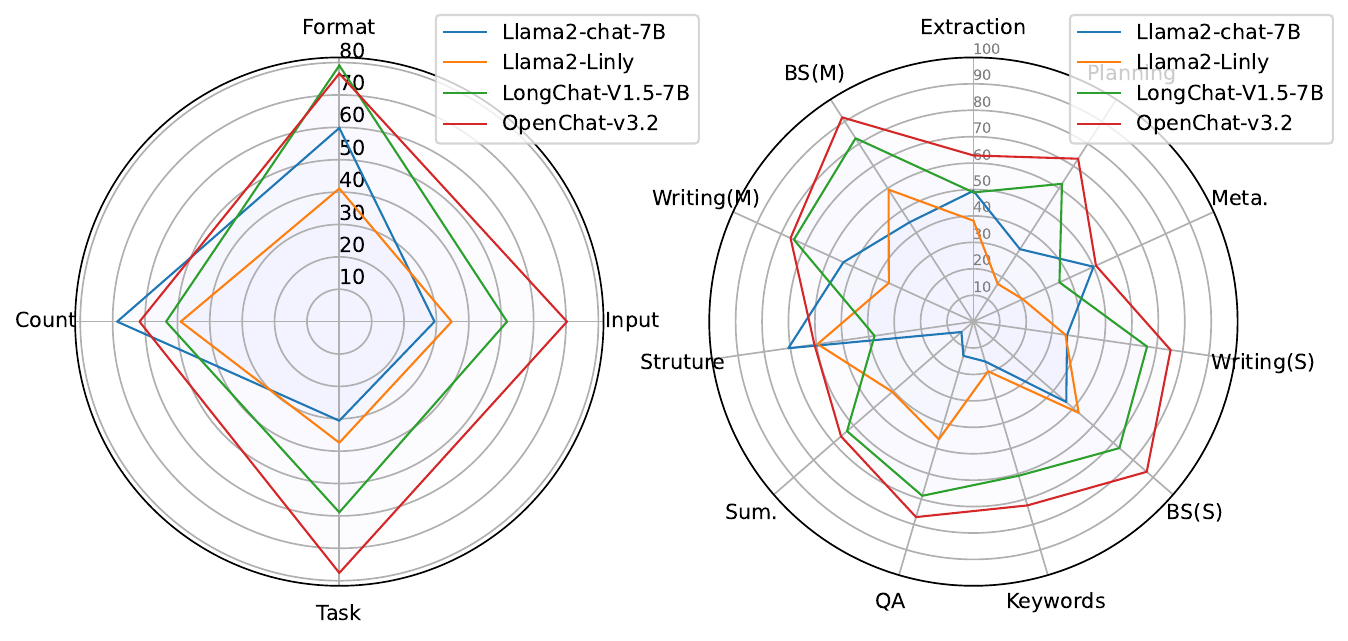} 
    \caption{The performance of LLMs grounded on the same base model~\cite{touvron2023llama} regarding different tasks and criteria.}
    \captionsetup{font={small}} 
    \label{fig:040visual}
\end{figure}

%% file: 050conclusion.tex
% In this work, we systematically investigate the complex understanding ability of LLMs.
% We propose a comprehensive set of features to identify complex instructions and a two-stage framework for dataset construction and finally construct a Chinese complex instruction evaluation dataset.
% Moreover, we design four evaluation criteria and corresponding metrics to assess LLMs' complex instruction understanding ability in a fine-grained and discriminative way.
% Furthermore, we conduct extensive experiments to compare the performance of representative models in our benchmark.

In this work, we systematically investigate the complex instructions following ability of LLMs.
We establish a framework comprising eight features for complex instructions, then construct an evaluation dataset covering nine tasks, and finally propose four evaluation criteria and corresponding metrics to assess LLMs' complex instruction understanding ability.
Furthermore, we conduct extensive experiments to compare the performance of representative models.

\section{Acknowledgements}
This work is supported by 
Science and Technology Commission of Shanghai Municipality Grant (No. 22511105902),
National Natural Science Foundation of China (No.62102095),
Shanghai Municipal Science and Technology Major Project (No.2021SHZDZX0103).
Yanghua Xiao is also a member of Research Group of Computational and AI Communication at Institute for Global Communications and Integrated Media, Fudan University.

%% file: 100appendix.tex
\clearpage
\appendix
\input{tables/100models}

\section{Data Evolution}
As introduced in the \textit{Data Evolution} part, we diversify the collected complex instructions through \textit{In-breadth Evolution} and complicate the simple instructions via \textit{In-depth Evolution}.
\textit{In-breadth Evolution} involves (1) Task Description Relocation, (2) Task Description Paraphrasing, and (3) Task Emulation, while \textit{In-depth Evolution} involves (4) Constraints Addition and (5) Multi-round Interaction.
Overall, we design several prompts to enhance the complexity and diversity of the data for various tasks.

\subsection{In-breadth Evolution}
We mainly design three prompts to diversify the data in Planning, QA, and Summarization tasks respectively.
\subsubsection{Planning}
\input{tables/100evo_planning}
\input{tables/100evo_planning_history}
We apply the \textit{Task Emulation} strategy when diversifying the data in the Planning task.
The prompts are shown in Tab.~\ref{tab:110evo_planning}, which mainly consists of two phases.
% In phase one, GPT-3.5-turbo需要根据用户输入的主题生成具体任务描述以及对应的工具描述，其中工具集描述包括工具的名字、简介以及需要的输入参数。
During phase one, GPT-3.5-turbo is required to generate specific \textit{Task Description} and corresponding \textit{Tools Descriptions} based on the theme provided by the user (e.g. \textit{maths} in the given example). 
The Tools Descriptions encompass each tool's name, a brief introduction, and the required input parameters.
During phase two, GPT-3.5-turbo is required to provide the planning process given the \textit{Task Description} and corresponding \textit{Tools Descriptions} generated in phase one.
The planning process consists of four main parts: the Task Description, Tools Descriptions, Output Format, and Histories.
An example of the Instruction generated from this two-phase prompt is shown in Tab.~\ref{tab:110evo_planning_history}.

% 值得注意的是，我们意识到GPT-3.5-turbo作为自动agent far from perfect. 为确保生成数据的质量，as shown in Tab.~\ref{tab:110evo_planning_history}，我们通过人工填入工具的正确的返回值，从而确保histories中的推理过程和结果均为正确的。
It is worth noting that we acknowledge GPT-3.5-turbo is far from a perfect automated agent~\cite{liu2023agentbench}. 
In order to ensure the quality of the generated data, as depicted in Table~\ref{tab:110evo_planning_history}, we manually enter the correct return values of the tool to ensure that both the planning process and results in the histories are accurate.

\subsubsection{Summarization}
\input{tables/100evo_sum}
The prompt we use to diversify the data in the \textit{Summarization} task is shown in Tab.~\ref{tab:100evo_sum}.
% 我们提供了设计task description的不同原则，最终通过Task Description Relocation以及Task Description Paraphrasing策略为大量的输入文本设计了不同的task description。
We present various underlying principles for designing task descriptions for  \textit{Summarization} task in our prompt.
These principles mainly employ the \textit{Task Description Relocation} and \textit{Task Description Paraphrasing} strategies.
We finally generate task descriptions for a total of 100 input text provided.

\subsubsection{QA}
\input{tables/100evo_QA}

The prompt utilized to diversify the data in the \textit{QA} task is shown in Tab.~\ref{tab:100evo_QA}.
% 对于给定的输入文本，我们要求模型以不同的方式提出更多样的问题，从而diversify了task description。在这里，我们的prompt也主要利用了\textit{Task Description Relocation} and \textit{Task Description Paraphrasing} strategies
In order to enhance the diversity of task descriptions, we require the model to generate a wider range of questions when provided with a given input text. 
Here, our prompt primarily employs strategies such as \textit{Task Description Relocation} and \textit{Task Description Paraphrasing}.

\subsection{In-depth Evolution}
We design two prompts to complicate the simple instructions collected regrading the \textit{Well-guided Writing} and \textit{Brainstorming }task. 
Both prompts utilize the \textit{Constraints Addition} and \textit{Multi-round Interaction }strategies.

\subsubsection{Well-guided Writing}
\input{tables/100evo_writing}
The prompt to increase the complexity of the basic instruction in the \textit{Well-guided Writing} task can be seen in Tab.~\ref{tab:100evo_writing}.
In order to simulate human-like multi-round modifications during the writing process, we define three atomic operations:
(1) \textit{Count Limit} establishes clear requirements for word or sentence count. 
(2) \textit{Specification} involves specifying crucial details such as keywords, hashtags, and URLs to ensure precise alignment with specific needs. 
(3) \textit{Revision} involves proposing dynamic and objective amendments to enhance the writing style.
By employing these operations, the requirements can be more specific, leading to more effective guidance for the generated results. 
We ensure that any modifications introduced are objective and can be evaluated automatically.
These atomic operations can be reused during the composition process.
\subsubsection{Brainstorming}
\input{tables/100evo_BS}
The prompt that we design for enhancing the complexity of simple instruction in the \textit{Brainstorming} task is shown in Tab.~\ref{tab:100evo_BS}
We define two atomic operations to mimic the human thinking process: 
(1) \textit{Modification} includes altering the output format such as JSON, XML, CSV, Markdown table, Python list, numeric sequence, etc. Additionally, word, sentence, or sample count limits can be imposed. Key information like keywords, hashtags, URLs, and language can also be incorporated into the instruction. 
(2) \textit{Specification} Further inquire about the specific details or ask for more information.
The GPT-3.5-turbo can simulate human thought processes by combining the two atomic operations.
The history of multiple calls to these operations can be aggregated into multi-turn dialogues.
The final evolved instructions shown in the prompt can serve as complex single-turn instructions, challenging the model to accomplish multiple tasks within a single round of instruction.

\section{Scoring Keywords Annotation}
We propose four criteria for complex instruction understanding, namely Count Limit, Answer Format, Task-prescribed phrases, and Input-dependent query, as introduced in our evaluation system.
% Among them, the latter three all 涉及到了scoring keywords的标注。对于Answer Format，keywords是客观的（例如JSON格式的keywords是``{'',``}'',``"''），因此人类直接标注。对于Task-prescribed phrases, and Input-dependent query，我们采用GPT4和人类协作的方式。对于Task-prescribed phrases，我们要求GPT4直接从Task description中抽取和任务目标有关的关键phrases，包含keyword、predefined functions等。对于Input-dependent query，我们要求GPT4先回答Instruction，然后要求从它的答案中总结和Input text相关的部分。最终，三个标注者要求对GPT4的标注结果进行检查和补充，只有两个以上的标注者均cover的关键词能进入最终的label set.
mong these criteria, the latter three involve the annotation of scoring keywords. 
For Answer Format, objective keywords such as ``\{", and ``\}" are directly annotated by humans. 
For Task-prescribed phrases and Input-dependent query, we employ a collaborative approach with GPT4 and humans. 
For Task-prescribed phrases, we require GPT4 to extract key phrases related to the task objective directly from the task description, such as keywords and predefined functions. 
For Input-dependent query, we ask GPT4 to answer the instruction first and then summarize the keywords of its answer that are relevant to the input text. 
Finally, the annotations by three evaluators are checked and supplemented, and only keywords covered by two or more evaluators are included in the final label set.

\section{Models}
% 我们将评估的模型具体信息展示在表x。我们评估了18个Chinese-oriented Models以及15个English-oriented Models，它们的区别在于它们预训练语料库的中文数据比例是多还是少。其中，Chinese-oriented Models又根据模型是重头训练（From Scratch，FS）还是接着English-oriented Models继续训练（Continue Pretraining, CP）。我们展示了它们的base model、模型大小、支持的上下文长度、instruction tuning阶段用的样本数量、是否经过RLHF，以及Chinese-oriented Model (CP)是否扩充了词表中的中文字符。
We present the details of our evaluated models in Table~\ref{tab:100models}.
Overall, we evaluate \modelNumch Chinese-oriented models and \modelNumeg English-oriented models.
The difference between Chinese-oriented models and English-oriented models lie in the proportion of Chinese data in their pretraining corpus.
Among them, Chinese-oriented models are further categorized based on whether they are trained from scratch (From scratch, FS) or continue pretraining from English-oriented models (Continue Pretraining, CP).
We provide details on their base model, model size, supported context length, the number of samples used in the instruction tuning phase, whether they adopt reinforcement learning with human feedback, and whether the Chinese-oriented model (CP) has expanded the Chinese characters in its vocabulary.

\footnotetext[1]{https://huggingface.co/Qwen/Qwen-7B}
\footnotetext[2]{https://huggingface.co/baichuan-inc/Baichuan-13B-Chat}
\footnotetext[3]{https://huggingface.co/Abbey4799/kw-cutegpt-13b-ift-lora}
\footnotetext[4]{https://huggingface.co/LinkSoul/Chinese-Llama-2-7b}
\footnotetext[5]{https://huggingface.co/FlagAlpha/Llama2-Chinese-7b-Chat}
\footnotetext[6]{https://huggingface.co/Linly-AI/Chinese-LLaMA-2-7B-hf}
\footnotetext[7]{https://huggingface.co/OpenBuddy/openbuddy-llama2-13b-v8.1-fp16}

%% file: tables/100models.tex
\setlength\tabcolsep{4pt}
\begin{table*}[t]
  \centering
  \resizebox{\textwidth}{!}{
\begin{tabular}{ccccccc}
    \toprule
\textbf{Model}                                 & \textbf{Base Model} & \textbf{Size} & \textbf{\begin{tabular}[c]{@{}c@{}}Vocabulary \\ Expansion\end{tabular}} & \textbf{\begin{tabular}[c]{@{}c@{}}Supported \\ Context \\ Length\end{tabular}} & \textbf{\begin{tabular}[c]{@{}c@{}}\# IFT \\ samples\end{tabular}} & \textbf{RLHF}             \\
    \midrule
    \rowcolor[gray]{0.95}\multicolumn{7}{c}{\textit{Chinese-oriented Models (From Scratch)}} \\
InternLM~\cite{2023internlm}                         & InternLM-chat-7B   & 7B          & N/A          & 8k            & 500w        & T  \\
BatGPT-sirius~\cite{li2023batgpt}                & BatGPT             & 15B         & N/A          & 32k           & \checkmark         & T  \\
Qwen\footnotemark[1]                                                 & Qwen-7B            & 7B          & N/A          & 8k            & \checkmark        & F  \\    
Baichuan-chat\footnotemark[2]                                        & Baichuan-Base      & 13B         & N/A          & 4k            & \checkmark         & F  \\   
MOSS (moss-moon-003-sft)~\cite{sun2023moss}                              & moss-moon-003-base & 16B         & N/A          & 2k            & 110w        & F  \\  
ChatGLM~\cite{zeng2023glm-130b}                      & ChatGLM-6B         & 6B          & N/A          & 2k            & \checkmark        & T  \\   
ChatGLM2~\cite{zeng2023glm-130b}                     & ChatGLM-6B         & 6B          & N/A          & 8k            & \checkmark         & T  \\    
ChatGLM2-32k~\cite{zeng2023glm-130b}                 & ChatGLM-6B         & 6B          & N/A          & 32k           & \checkmark          & T  \\
    \midrule
    \rowcolor[gray]{0.95}\multicolumn{7}{c}{\textit{Chinese-oriented Models (Continue Pretraining)}} \\
Baize-V2~\cite{xu2023baize}                             & Llama1             & 7B, 13B      & F            & 2k            & 5w        & F  \\
BELLE~\cite{BELLE}                             & BLOOMZ-7B1-mt             & 7B      & F            &      1k       &     200w    &   F \\
Chinese-Alpaca-V1~\cite{Chinese-LLaMA-Alpaca}        & Llama1             & 7B, 13B, 33B  & T            &  8k           & 200w, 300w, 430w  & F  \\
CuteGPT\footnotemark[3]                                             & Llama1             & 13B         & T            & 2k            & 110w       & F  \\
Llama2-LinkSoul\footnotemark[4] & Llama2             & 7B          & F            & 4k            & 1000w      & F  \\
Llama2-FlagAlpha\footnotemark[5]     & Llama2             & 7B          & F            & 4k          & \checkmark         & F  \\
Llama2-Linly\footnotemark[6]      & Llama2             & 7B          & T            & 4k            & 120w       & F  \\
Llama2-OpenBuddy\footnotemark[7]  & Llama2             & 13B         & T            & 4k            & 100w       & F \\
    \midrule
    \rowcolor[gray]{0.95}\multicolumn{7}{c}{\textit{English-oriented Models}} \\                                       
Llama2-chat~\cite{touvron2023llama}                                          & Llama2             & 7B, 13B, 70B         & N/A            & 4k            & 10w       & T     \\
Vicuna-V1.3~\cite{zheng2023judging}                       & Llama1             & 7B, 13B, 33B  & N/A             & 2k            & 12w         & F     \\
Vicuna-V1.5~\cite{zheng2023judging}                    & Llama2             & 7B, 13B      & N/A             & 16k            & 12w         & F    \\
WizardLM~\cite{xu2023wizardlm}                       & Llama1             & 13B         & N/A             & 2k            & 25w      & F     \\
LongChat-V1~\cite{longchat2023}                                          & Llama1             & 7B, 13B         & N/A             & 16k           &   8w, 2w         & F    \\
LongChat-V1.5~\cite{longchat2023}                                        & Llama2             & 7B         & N/A             & 32k           & \checkmark       & F     \\
OpenChat-V3.2~\cite{openchat}                                        & Llama2             & 13B         & N/A             & 4k            & 0.6w   & F    \\

  \cdashlinelr{1-7}

GPT-3.5-turbo                                        & -             & -         & N/A            & 16k            &  \checkmark   & T    \\
GPT-4                                        &  -             & -         & N/A           & 16k            & \checkmark  & T    \\
    \bottomrule
\end{tabular}
}
    % 此处'-'以及\checkmark代表细节未公开。Vocabulary Expansion代表继续预训练的Chinese-oriented Models是否在词表中扩充了中文字符。supported context length是指模型支持的上下文长度。# IFT samples是指instruction tuning阶段训练使用的样本数量。RLHF列的T和F代表模型是否采用reinforcement with human feedback。
  \caption{Models evaluated in this paper. 
  The symbols '-' and \checkmark denote that details are undisclosed.
  \textit{Vocabulary Expansion} indicates whether Chinese-oriented Models (Continue Pretraining) have expanded their vocabulary to include Chinese characters.
  \textit{\# IFT samples} denotes the number of samples used in the instruction tuning phase. 
  The \textit{RLHF} column indicates whether the model adopts reinforcement learning with human feedback. }
  \label{tab:100models}
\end{table*}

%% file: tables/100evo_planning.tex
\begin{table*}[t]
\small
  \centering
    \begin{tabularx}{\linewidth}{X}
    \toprule
    \rowcolor[gray]{0.95}\multicolumn{1}{c}{\textbf{I: Task \& Tools Descriptions Generation}} \\
    \makecell[l]{
    \color{gray}{/* \textit{Task prompt} */}\\
    Suppose you're a good planner for designing complex planning tasks in \color{purple}{\textbf{maths}} \color{black}{and provide some implicitly useful} tools for solving the \\ problem. 
    Your task is to design tasks that need multi-step operations and thoughts and design tools that can help users to solve the problem. \\
    \color{gray}{/* \textit{Output Format} */}\\
    You should return the answer in the format as described \\
    \{ 
    ``task": ``\textless a brief task description\textgreater", \\
    \quad ``tools": [ \{ ``name": ``\textless tool name\textgreater", ``description": ``\textless tool description\textgreater", ``input": \{ ``\textless name \textgreater": ``\textless value \textgreater", ... \}\}, ... ] 
    \} \\
    \color{gray}{/* \textit{Example} */}\\
    For example:
    \{ 
    ``\textbf{Task}": ``\color[rgb]{0.81, 0.41, 0.22}{\underline{You are an AI that helps users book flights. Ask the user for their travel plans, then show them flights,}}\\
        \qquad\quad\qquad\quad  \qquad\quad \quad  \color[rgb]{0.81, 0.41, 0.22}{\underline{and book the flights they select.}}", \\
    \qquad\quad\qquad\quad ``\textbf{Tools}": [ 
    \{ 
    \color[rgb]{0,0.39,0}{\underline{``name": ``findFlights",} \underline{``description": ``searches for available flights", }}\\
    \qquad\quad\qquad \qquad\quad\qquad\quad \color[rgb]{0,0.39,0}{\underline{``input": \{ } \underline{``Origin": ``\textless airport code\textgreater",} \underline{``Destination": ``\textless airport code\textgreater",} \underline{``DepartureDate": ``\textless date\textgreater",}} \\
    \qquad\quad\qquad \qquad\qquad\qquad\qquad \color[rgb]{0,0.39,0}{\underline{``ReturnDate": ``\textless date\textgreater",} \underline{``Passengers": ``\textless count\textgreater"}} 
    \} 
    \}, 
    .. ] 
    \}}\\
    \midrule
    \rowcolor[gray]{0.95}\multicolumn{1}{c}{\textbf{II: Planning Process Generation}} \\
    \makecell[l]{\color{gray}{/* \textit{Task Description} */}\\
    \color[rgb]{0.81, 0.41, 0.22}{\underline{[\textit{Task Description} from Phase 1]}}. \\
    \color{gray}{/* \textit{Tools Descriptions} */}\\
    \color[rgb]{0,0.39,0}{\underline{[\textit{Tools Descriptions} from Phase 1]}}. \\
         \color{gray}{/* \textit{Output Format} */}\\
    You should only respond in JSON format as described below \\
    Response Format: \\
    \{ \\
    \quad \{ ``thoughts": \{ \\
    \quad \quad \quad ``thought": ``\textless your current thought\textgreater'', \\
    \quad \quad \quad ``reasoning": ``\textless self reflect on why you made this decision'', \\
    \quad \quad \quad "plan": ``short bulleted list that conveys long-term plan'' \\
    \quad \quad \}, \\
    \quad \quad``command": \{ \\
    \quad \quad \quad ``name": ``command name'', \\
    \quad \quad \quad ``input": \{ \\
    \quad \quad \quad \quad ``\textless name\textgreater'': ``\textless value\textgreater'' \\
    \quad \quad \quad \} \\
    \quad \quad \}, \\
    \} \\
    Ensure the response can be parsed by Python \texttt{json.loads} \\
    \color{gray}{/* \textit{Histories} */}\\
    And then the system will execute the command and give you the 
    result and log the execution history below.  \\
    Please mind the history and the given result.\\
    \\
    System: This reminds you of these events from your past:\\
    \color{purple}{\textbf{[\textit{History}]}}\\
    Human: Stay focused on the history and determine which next 
    command to use, and respond using the format specified above:}\\
    \bottomrule
    \end{tabularx}
  \caption{
  % 基于Task Emulation strategy多样化Planning任务数据的prompts，during Data evolution process. Overall，data evolution for Planning task包含两个阶段：生成Task & Tools Descriptionsd以及Planning Process Genertion. 需要人工填写的部分are \color{purple}{highlighted}.
  The prompts for diversifying the data in the \textit{Planning} task during the \textit{Data Evolution} process.
  Overall, the data evolution for the \textit{Planning} task consists of two phases: Tools \& Task Description Generation and Planning Process Generation. 
  The information that requires manual input is \color{purple}{\textit{highlighted}}.
  \color{black}{An example of the \textit{Instruction} generated from this two-phase prompt is shown in Tab.~\ref{tab:110evo_planning_history}.}
  }
  \label{tab:110evo_planning}
\end{table*}

% \begin{table}[ht]
% \centering
% \caption{Task and Tools}
% \begin{tabularx}{\textwidth}{|X|}
% \hline
% \textbf{Data} \\
% \hline
% \{ \\
% \quad "task": "<a brief task description>", \\
% \quad "tools": [ \\
% \qquad \{ \\
% \qquad\quad "name": "<tool name>", \\
% \qquad\quad "description": "<tool description>", \\
% \qquad\quad "input": \{ \\
% \qquad\qquad "<name>": "<value>", \\
% \qquad\qquad ... \\
% \qquad\quad \}, \\
% \qquad \}, \\
% \qquad ... \\
% \quad ] \\
% \}
% \\
% \hline
% \end{tabularx}
% \end{table}

%% file: tables/100evo_planning_history.tex
\begin{table*}[t]
\small
  \centering
    \begin{tabularx}{\linewidth}{X}
    \toprule
    \makecell[l]{
    \color{gray}{/* \textit{Task Description} */}\\
\color[rgb]{0.81, 0.41, 0.22}{Design a task to find the area of a triangle and provide tools to assist with the calculations.} \\
     \color{gray}{/* \textit{Tools Descriptions} */}\\
Tools: {[}\\
\quad \color[rgb]{0,0.39,0}{\{``name": ``\textbf{calculateSemiPerimeter}", ``description": ``calculates the semi-perimeter of the triangle", }\\
\quad \quad \color[rgb]{0,0.39,0}{``input": \{ ``sideA": ``\textless length of side A\textgreater", ``sideB": ``\textless length of side B\textgreater", ``sideC": ``\textless length of side C\textgreater"\}\}, }\\
\quad \color[rgb]{0,0.39,0}{\{ ``name": ``\textbf{displayArea}", ``description": ``displays the calculated area of the triangle to the user", }\\
\quad \quad \color[rgb]{0,0.39,0}{``input": \{ ``area": ``\textless calculated area\textgreater"\}\},}\\
\quad \color[rgb]{0,0.39,0}{\{ ``name": ``\textbf{calculateArea}", }\\
\quad \quad \color[rgb]{0,0.39,0}{``description": ``calculates the area of the triangle using Heron's formula",} \\
\quad \quad \color[rgb]{0,0.39,0}{``input": \{ ``\textbf{semiPerimeter}": ``\textless semi-perimeter of the triangle\textgreater", ``sideA": ``\textless length of side A\textgreater", ``sideB": ``\textless length of side B\textgreater", }\\
\quad \quad \quad \color[rgb]{0,0.39,0}{``sideC": ``\textless length of side C\textgreater"  \}\}, }\\
\quad \color[rgb]{0,0.39,0}{\{  ``name": ``\textbf{inputTriangleSides}", ``description": ``asks the user for the lengths of the three sides of the triangle", ``input": \{\}\}}\\
{]} \\
     \color{gray}{/* \textit{Output Format} */}\\
You should only respond in JSON format as described below \\
Response Format: \\
\{ \\
\quad \{ ``thoughts": \{ \\
\quad \quad \quad ``thought": ``\textless your current thought\textgreater'', \\
\quad \quad \quad ``reasoning": ``\textless self reflect on why you made this decision\textgreater'', \\
\quad \quad \quad ``plan": ``short bulleted list that conveys long-term plan'' \\
\quad \quad \}, \\
\quad \quad``command": \{ \\
\quad \quad \quad ``name": ``command name'', \\
\quad \quad \quad ``input": \{ \\
\quad \quad \quad \quad ``\textless name\textgreater'': ``\textless value\textgreater'' \\
\quad \quad \quad \} \\
\quad \quad \}, \\
\} \\
Ensure the response can be parsed by Python \texttt{json.loads} \\
\color{gray}{/* \textit{Histories} */}\\
System: This reminds you of these events from your past: \\
{[} \\
\quad \{ ``Reply": \\
\quad \quad \quad  \{ \\
\quad \quad \quad \quad   ``thoughts": \{ \\
\quad \quad \quad \quad \quad   ``thought":  ``I can start by asking for the lengths of the three sides of the triangle using the `inputTriangleSides' command.", \\
\quad \quad \quad \quad \quad   ``reasoning":  ``By asking for the lengths of the sides, I can obtain the necessary input to calculate the area of the triangle.", \\
\quad \quad \quad \quad \quad   ``plan": {[} \\
\quad \quad \quad \quad \quad \quad ``Ask the user for the lengths of the three sides of the triangle using the `inputTriangleSides' command. " \\
\quad \quad \quad \quad  \quad {]}, \\
\quad \quad \quad \quad  \quad   ``command": \{ ``name":  ``inputTriangleSides", ``input": \{\}\} \\
\quad \quad \quad \quad   \}, \\
\quad \quad \quad  \}, \\
\quad \quad ``Result": \color{purple}{``\textbf{I have a triangle with side lengths 5, 12, 13}}" \\
\quad \}, \\
\quad \{ \\
\quad \quad ``Reply": \\
\quad \quad \quad  \{ \\
\quad \quad \quad \quad   ``thoughts": \{ \\
\quad \quad \quad \quad \quad   ``thought":  ``I can calculate the semi-perimeter of the triangle using the `calculateSemiPerimeter' command.", \\
\quad \quad \quad \quad \quad   ``reasoning":  ``By calculating the semi-perimeter, I can then use it to calculate the area of the triangle using Heron's formula.", \\
\quad \quad \quad \quad \quad   ``plan": {[} \\
\quad \quad \quad \quad \quad \quad"Calculate the semi-perimeter of the triangle using the `calculateSemiPerimeter' command. " \\
\quad \quad \quad \quad  \quad {]}, \\
\quad \quad \quad \quad  \quad    ``command": \{ ``name":  ``calculateSemiPerimeter", ``input": \{ ``sideA": 5, ``sideB": 12, ``sideC": 13\}\} \\
\quad \quad \quad \quad   \}, \\
\quad \quad \quad  \}, \\
\quad \quad ``Result": \color{purple}{``\textbf{15}''} \\
\quad \} \\
{]} \\
Human: Determine which next command to use, and respond using the format specified above: \\
    }\\
    \bottomrule
    \end{tabularx}
  \caption{
  % 基于Tab.~\ref{tab:110evo_planning}的two-phase prompts得到的新的Task description for Planning task。
  The newly generated \textit{Instruction} for the \textit{Planning} task during data evolution, derived from the two-phase prompts in Tab.~\ref{tab:110evo_planning}.
  The information that requires manual input is \color{purple}{\textit{highlighted}}.
  }
  \label{tab:110evo_planning_history}
\end{table*}

%% file: tables/100evo_sum.tex
% \begin{table}[t]
% \small
%   \centering
%     \begin{tabularx}{\linewidth}{X}
%     \toprule
%     \makecell[l]{
%     You are a question-generation agent that can pose multiple \\
%     questions in line with a given text description, and these \\
%     questions should also have a certain level of difficulty. \\
%     Based on the provided text, pose questions that align with \\
%     its description. The answers to the questions should be found \\ 
%     within the text, and they shouldn't be explicitly stated; \\
%     Instead, they should require inference to deduce. \\
%     }\\
%     \bottomrule
%     \end{tabularx}
%   \caption{
%   % 基于Tab.~\ref{tab:110evo_planning}的two-phase prompts得到的新的Task description for Planning task。
%   The newly generated \textit{Instruction} for the \textit{Planning} task during data evolution, derived from the two-phase prompts in Tab.~\ref{tab:110evo_planning}.
%   The information that requires manual input is \color{purple}{\textit{highlighted}}.
%   }
%   \label{tab:110evo_QA}
% \end{table}

\begin{table*}[t]
\small
  \centering
    \begin{tabularx}{\linewidth}{X}
    \toprule
    \makecell[l]{
    You are a task generator, and your role is to create a task description to describe the task of \textit{summarizing customer service conversations}.\\
    You can generate the following task descriptions:\\
    1. Given the conversation records between the customer service agent (A) and the user (Q), please \textit{summarize the content of the dialogue}  \\
    \quad and \textit{list the main points}.\\
    2. \textit{Summarize the key information} in the conversation records between customer service agent (A) and the user (Q).\\
    3. For the provided conversation records between the customer service agent (A) and the user (Q), \textit{summarize the dialogue content} and \\
    \quad  \textit{list the main points}. \textit{Describe the issues and solutions} between the customer service agent and the user, including the user's questions, \\
    \quad  the agent's answers, and the solutions. At the same time, summarize the key information from the conversation records.\\
    4. Please analyze and summarize the provided conversation records between the customer service agent (A) and the user (Q), \\
    \quad  \textit{describe the issues} raised by the user, and \textit{the agent's responses and solutions}, and identify the \textit{key information} in the dialogue.\\
    5. Based on the conversation records between the customer service agent (A) and the user (Q), \textit{organize the main content} of the dialogue \\
    \quad  and \textit{summarize the key information and solutions}.
    }\\
    \bottomrule
    \end{tabularx}
  \caption{
  % 基于Tab.~\ref{tab:110evo_planning}的two-phase prompts得到的新的Task description for Planning task。
  The prompts for diversifying the data in the \textit{Summarization} task during the \textit{Data Evolution} process.
  }
  \label{tab:100evo_sum}
\end{table*}

%% file: tables/100evo_QA.tex
\begin{table*}[t]
\small
  \centering
    \begin{tabularx}{\linewidth}{X}
    \toprule
    \makecell[l]{
    You are a question-generation agent that can pose multiple 
    questions in line with a given text description, and these 
    questions should also  \\
    have \textit{a certain level of difficulty}. 
    Based on the provided text, pose questions that align with 
    its description. The answers to the questions \\
    should be \textit{found within the text}, and they \textit{shouldn't be explicitly stated}; 
    Instead, they should \textit{require inference to deduce}. 
    }\\
    \bottomrule
    \end{tabularx}
  \caption{
  % 基于Tab.~\ref{tab:110evo_planning}的two-phase prompts得到的新的Task description for Planning task。
   The prompts for diversifying the data in the \textit{QA} task during the \textit{Data Evolution} process.
  }
  \label{tab:100evo_QA}
\end{table*}

%% file: tables/100evo_writing.tex
\begin{table*}[t]
\small
  \centering
    \begin{tabularx}{\linewidth}{X}
    \toprule
    \makecell[l]{
      \color{gray}{/* \textit{Task Prompt} */}\\
As a skilled writer, your objective is to effectively achieve a simple writing goal by implementing the following strategies:\\
1. \textit{Precisely Define Requirements}: Continuously elevate the accuracy and specificity of your requirements to effectively guide \\ 
\quad the generated results. \\
2. \textit{Objective Revisions}: When introducing modifications, ensure that they are objective and amenable to automated evaluation. \\
\quad Avoid subjective and vague instructions, to maintain a consistent and coherent tone.\\
      \color{gray}{/* \textit{Defined Atomic Operations} */}\\
Additionally, you have the flexibility to combine various operations to fine-tune the output:\\
1.``\textbf{Count Limit}": Establish clear word or sentence count requirements, allowing you to strike the right balance between conciseness \\
\quad and comprehensiveness. \\
2.``\textbf{Specification}": Specify crucial details like keywords, hashtags, and URLs to align the writing precisely with your specific needs. \\
3.``\textbf{Revision}": Propose dynamic and objective amendments to enhance the writing style.\\
\quad By following these guidelines, you can harness the full potential of AI-generated content and accomplish your writing objectives with \\
\quad precision and excellence.\\
    \color{gray}{/* \textit{Output Format} */}\\
To fulfill this task, you are expected to provide your responses in the following JSON format: \\
\{ \\
\quad ``Operations": [ \\
\quad \quad \{ \\
\quad \quad \quad ``operation": \textless ``Count limit", ``Specification" or ``Revision"\textgreater, \\
\quad \quad \quad ``thoughts": \textless Your thinking process\textgreater, \\
\quad \quad \quad ``takeways": \textless Briefly summarize your thought process into a short instruction\textgreater \\
\quad \quad \} \\
\quad ] \\
\} \\
    }\\
    \color{gray}{/* \textit{Histories} */}\\
\textbf{Input:} \\
\quad Create a summary for a given article. [An article] \\
\textbf{Output:} \\
\{ \\
\quad ``Operations": [ \\
\quad \quad \{ \\
\quad \quad \quad ``operation": ``\textbf{Count limit}", \\
\quad \quad \quad ``thoughts": ``I'd like the summary to be neither too concise nor excessively lengthy, so I'd prefer to limit it to three sentences.", \\
\quad \quad \quad ``takeways": ``Limit the length to three sentences." \\
\quad \quad \}, \\
\quad \quad \{ \\
\quad \quad \quad ``operation": ``\textbf{Revision}", \\
\quad \quad \quad ``thoughts": ``The response might be too short and plain.", \\
\quad \quad \quad ``takeways": ``The response could benefit from a touch of eloquence." \\
\quad \quad \}, \\
\quad \quad \{ \\
\quad \quad \quad ``operation": ``\textbf{Specification}", \\
\quad \quad \quad ``thoughts": ``I should define a set of keywords that can better guide the summary.", \\
\quad \quad \quad ``takeways": ``Requesting retention of keywords: wildflowers, summer." \\
\quad \quad \} \\
\quad ] \\
    \color{gray}{/* \textit{Input} */}\\
\textbf{Input:} Craft an Instagram post caption for a photo of my dog and me playing at the beach.
\} \\
    \bottomrule
    \end{tabularx}
  \caption{
  % 基于Tab.~\ref{tab:110evo_planning}的two-phase prompts得到的新的Task description for Planning task。
    The prompt for enhancing the complexity of the simple instruction in the \textit{Well-guided Writing} task during the \textit{Data Evolution} process. 
    % 我们定义了三个原子操作，让GPT-3.5-turbo通过组合这三个原子操作模拟人对well-guided写作更细粒度的多轮修改，原子操作可重复使用。
    Three atomic operations have been specifically defined to facilitate GPT-3.5-turbo in its ability to simulate human-like multi-round modifications during the writing process. 
    These atomic operations can be reused.
  }
  \label{tab:100evo_writing}
\end{table*}

%% file: tables/100evo_BS.tex
\begin{table*}[t]
\small
  \centering
    \begin{tabularx}{\linewidth}{X}
    \toprule
    \makecell[l]{
      \color{gray}{/* \textit{Task Prompt} */}\\
As a thinker, when presented with a simple thinking problem, your objective is to simulate human thinking, following these steps:\\
1. \textit{Refine the requirements} of the thinking questions to render the results more specific, intuitive, easily consultable and comprehensible.\\
2. \textit{Engage in multiple rounds} of dialogue to continually probe and gain insights into the issue.\\
\color{gray}{/* \textit{Defined Atomic Operations} */}\\
You can combine the following operations:\\
1. ``\textbf{Modification}": Add, delete, modify the restrictions of the Evolved Instruction, including its output format (JSON, XML, CSV, 
 \\
 \quad Markdown table, Python list, Numeric sequence, etc.), imposing word/sentence/sample count limits, and incorporating key information \\
  \quad(keywords, hashtags, URLs, etc.), language.\\
2. ``\textbf{Specification}": Further inquire about the specific details or ask for more information.\\
    \color{gray}{/* \textit{Output Format} */}\\
To fulfill this task, you are expected to provide your responses in the following JSON format: \\
\{ \\
\quad ``Operations": [ \\
\quad \quad \{ \\
\quad \quad \quad ``operation": \textless ``Modification" or ``Specification"\textgreater, \\
\quad \quad \quad ``thoughts": \textless Your thinking process\textgreater, \\
\quad \quad \quad ``takeways": \textless Briefly summarize your thought process into a short instruction\textgreater \\
\quad \quad \quad ``evolved instruction": \textless A more complex instruction according to your selected operation\textgreater \\
\quad \quad \} \\
\quad ] \\
\} \\
    }\\
    \color{gray}{/* \textit{Histories} */}\\
\textbf{Input:} \\
\quad Provide five innovative or improved methods to solve everyday life problems. \\
\textbf{Output:} \\
\{ \\
\quad ``Operations": [ \\
\quad \quad \{ \\
\quad \quad \quad ``operation": ``\textbf{Modification}", \\
\quad \quad \quad ``thoughts": ``For easier readability, I'd like the output in the form of a Markdown table. Specifically, I'm interested in keywords, summaries, and steps for each method.", \\
\quad \quad \quad ``takeways": {[}``Output in Markdown table format", ``Including keywords, summaries, and steps"{]} \\
\quad \quad \quad ``evolved instruction": {[}``Present five innovative or improved methods for solving everyday life problems through Markdown table format, including keywords, introductions, and steps."{]} \\
\quad \quad \}, \\
\quad \quad \{ \\
\quad \quad \quad ``operation": ``\textbf{Modification}", \\
\quad \quad \quad ``thoughts": `` The English version would be more convenient for me to read.", \\
\quad \quad \quad ``takeways": {[}``Translate into English."{]} \\
\quad \quad \quad ``evolved instruction": {[}``In Markdown table format, present five innovative or improved methods for solving everyday life problems, including keywords, summaries, and steps, and then translate into English."{]} \\
\quad \quad \} \\
\quad ] \\
    \color{gray}{/* \textit{Input} */}\\
\textbf{Input:} List three animals of different species. \\
\} \\
    \bottomrule
    \end{tabularx}
  \caption{
  % 基于Tab.~\ref{tab:110evo_planning}的two-phase prompts得到的新的Task description for Planning task。
    The prompt for enhancing the complexity of the simple instruction in the \textit{Brainstorming} task during the \textit{Data Evolution} process. 
    % GPT-3.5-turbo可以组合我们定义的三个原子操作模拟人的思考过程。
  }
  \label{tab:100evo_BS}
\end{table*}